\PassOptionsToPackage{table,svgnames}{xcolor}
\RequirePackage{xcolor}

\documentclass[11pt,letterpaper]{mystyle}

\usepackage[all]{hypcap}
\usepackage{xcolor}
\usepackage[numbers,comma,compress]{natbib}
\usepackage{algorithm}
\usepackage{algorithmicx}
\usepackage{algpseudocode}
\usepackage{microtype}
\usepackage{graphicx}
\usepackage{multirow}
\usepackage{placeins}
\usepackage{tikz}
\tcbuselibrary{skins,breakable}
\usetikzlibrary{arrows.meta,positioning,shapes.geometric,shapes.symbols}
\usepackage{pgfplots}
\pgfplotsset{compat=1.18}
\pgfplotsset{
  raw-gated curve/.style={
    color=orange!85!black,
    line width=0.95pt,
    mark=diamond*,
    mark size=1.5pt,
    mark options={fill=white,draw=orange!85!black,line width=0.45pt}
  },
  calibrated-gated curve/.style={
    color=green!55!black,
    line width=1.05pt,
    mark=triangle*,
    mark size=1.6pt,
    mark options={fill=white,draw=green!55!black,line width=0.45pt}
  },
  baseline markers/.style={
    only marks,
    color=black!50,
    mark=square*,
    mark size=1.8pt,
    mark options={fill=black!35,draw=black!50,line width=0.35pt}
  },
  raw operating point/.style={
    only marks,
    color=orange!85!black,
    mark=*,
    mark size=2.25pt,
    mark options={fill=orange!45,draw=orange!85!black,line width=0.4pt}
  },
  calibrated operating point/.style={
    only marks,
    color=green!55!black,
    mark=*,
    mark size=2.25pt,
    mark options={fill=green!35,draw=green!55!black,line width=0.4pt}
  }
}
\expandafter\def\csname ver@subfig.sty\endcsname{}
\usepackage{booktabs} %
\usepackage{float}
\usepackage{bigstrut}

\usepackage{amsmath}
\usepackage{amssymb}
\usepackage{mathtools}
\usepackage{amsthm}
\usepackage{mathrsfs}
\usepackage{nicefrac}
\usepackage{dsfont}
\usepackage{enumitem}
\usepackage{subcaption}
\usepackage{graphicx,subfig}
\usepackage{cleveref}
\usepackage{bxcoloremoji}

\usepackage{float}

\usepackage[utf8]{inputenc} %
\usepackage[T1]{fontenc}    %
\usepackage{url}            %
\usepackage{booktabs}       %
\usepackage{amsfonts}       %
\usepackage{nicefrac}       %
\usepackage{microtype}      %
\usepackage{graphicx}
\usepackage{subcaption}
\usepackage{amssymb}
\usepackage{fdsymbol}
\usepackage{wrapfig}
\usepackage{lipsum}
\usepackage{enumitem}
\usepackage{stackengine}
\usepackage[font=small,labelfont=bf]{caption}
\usepackage{color}
\usepackage{adjustbox}

\usepackage{rotating}
\usepackage{makecell}

\newcommand{\x}{\mathbf{x}}

\definecolor{blanchedalmond}{rgb}{1.0, 0.92, 0.8}
\definecolor{carmine}{rgb}{0.59, 0.0, 0.09}
\definecolor{lightblue}{rgb}{0.22,0.45,0.70}%

\renewcommand{\mathbf}{\boldsymbol}

\makeatletter
\def\Ddots{\mathinner{\mkern1mu\raise\p@
\vbox{\kern7\p@\hbox{.}}\mkern2mu
\raise4\p@\hbox{.}\mkern2mu\raise7\p@\hbox{.}\mkern1mu}}
\makeatother

\definecolor{amaranth}{rgb}{0.9, 0.17, 0.31}
\definecolor{antiquebrass}{rgb}{0.8, 0.58, 0.46}
\definecolor{antiquefuchsia}{rgb}{0.57, 0.36, 0.51}
\definecolor{chromeyellow}{rgb}{0.31, 0.47, 0.26}

\newtcolorbox{AIbox}[2][]{aibox,title=#2,#1}
\definecolor{lightblue}{rgb}{0.22,0.45,0.70}%
\definecolor{Gray}{gray}{0.95}
\definecolor{Cornsilk}{rgb}{1.0, 0.97, 0.86}
\definecolor{RawAmber}{HTML}{F59E0B}
\colorlet{green}{CalTeal}
\colorlet{orange}{RawAmber}
\colorlet{blue}{ResaColor}
\colorlet{red}{CalRisk}

\usepackage{amsmath}

\usepackage[all]{hypcap}

\title{Evaluating Confidence-Gated Retrieval with Matched Trajectory Replay}

\runningtitle{Evaluating Confidence-Gated Retrieval with Matched Trajectory Replay}

\author[ ]{
  Prateek Chhikara  
}

\newcommand{\method}{\textsc{Calibrated-Gated}}
\newcommand{\rawmethod}{\textsc{Raw-Gated}}

\newcommand{\textittt}[1]{\texttt{#1}}
\newcommand{\mistrallogo}{\raisebox{-0.2\height}{\includegraphics[height=0.9em]{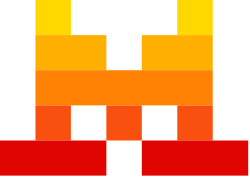}}}
\newcommand{\openailogo}{\raisebox{-0.2\height}{\includegraphics[height=0.9em]{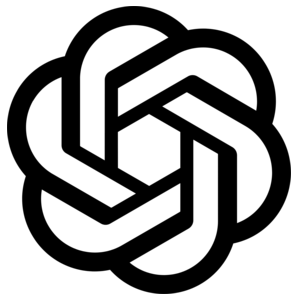}}}
\newcommand{\qwenlogo}{\raisebox{-0.2\height}{\includegraphics[height=0.9em]{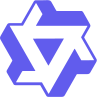}}}
\newcommand{\mistralfiglogo}{\raisebox{-0.25\height}{\includegraphics[height=0.8em]{logos/mistral-flat.png}}}
\newcommand{\openaifiglogo}{\raisebox{-0.25\height}{\includegraphics[height=0.8em]{logos/openai.png}}}
\newcommand{\qwenfiglogo}{\raisebox{-0.25\height}{\includegraphics[height=0.8em]{logos/qwen.png}}}
\newcommand{\reliabilitypanel}[2]{%
  \begin{minipage}[t]{0.315\linewidth}
    \centering
    \includegraphics[width=\linewidth,trim={0 0.45cm 0 0},clip]{#2}%
    \par\vspace{-0.1em}
    {\footnotesize #1}
  \end{minipage}%
}
\newcommand{\subplotgap}{\hspace{-0.04\linewidth}}
\newcommand{\figurelegend}{%
  {\scriptsize
  \begin{tabular}{@{}c@{\,}l@{\qquad}c@{\,}l@{\qquad}c@{\,}l@{}}
    \tikz[baseline=-0.5ex]{\draw[orange!85!black,line width=0.95pt] (0,0) -- (0.45,0); \draw[orange!85!black,fill=white,line width=0.45pt] plot[mark=diamond*,mark size=1.5pt] coordinates {(0.225,0)};} & Raw-gated &
    \tikz[baseline=-0.5ex]{\draw[green!55!black,line width=1.05pt] (0,0) -- (0.45,0); \draw[green!55!black,fill=white,line width=0.45pt] plot[mark=triangle*,mark size=1.6pt] coordinates {(0.225,0)};} & Calibrated-gated &
    \tikz[baseline=-0.5ex]{\draw[black!50,fill=black!35,line width=0.35pt] plot[only marks,mark=square*,mark size=1.8pt] coordinates {(0.225,0)};} & Baselines
  \end{tabular}}%
}

\newcommand{\bestval}[1]{\textbf{#1}}
\newcommand{\compareval}[1]{\cellcolor{green!18}{#1}}
\newenvironment{promptlist}{%
  \begin{list}{\texttt{-}}{%
    \setlength{\leftmargin}{1.8em}%
    \setlength{\labelwidth}{1em}%
    \setlength{\labelsep}{0.45em}%
    \setlength{\itemsep}{0.3em}%
    \setlength{\topsep}{0.2em}%
    \setlength{\parsep}{0pt}%
    \setlength{\partopsep}{0pt}%
  }%
}{%
  \end{list}%
}
\definecolor{promptfill}{HTML}{E8F1FF}
\definecolor{promptborder}{HTML}{0057B8}
\newtcolorbox{promptbox}[1][]{%
  enhanced,
  breakable,
  colback=promptfill,
  colframe=promptborder,
  boxrule=1pt,
  arc=4pt,
  left=12pt,
  right=12pt,
  top=10pt,
  bottom=10pt,
  before skip=10pt,
  after skip=10pt,
  overlay unbroken and first={},
  overlay middle={},
  overlay last={},
  overlay broken={}
}

\affil[ ]{University of Southern California, USA}

\correspondingauthor{Prateek Chhikara \href{mailto:pchhikar@usc.edu}{pchhikar@usc.edu}}

\begin{document}

\begin{abstract}
Interactive language-model agents use confidence signals to decide whether to answer immediately, retrieve additional evidence (from memory or external knowledge), or defer. Yet confidence is usually evaluated in isolation, without measuring the trajectory-level consequences of the actions it triggers.
We propose matched trajectory replay, a controlled protocol for comparing confidence-to-action mappings. The protocol holds candidate answer states, evidence points, budgets, and action costs fixed. We use it to compare raw verbalized confidence with post-hoc isotonic calibration in a multi-hop question-answering system using Mistral, GPT, and Qwen models on HotpotQA and MuSiQue datasets.
At the same numerical commitment threshold, calibration changes which questions agents ultimately commit to answering. Across all six model-dataset pairs, it increases accuracy among committed answers by up to 41 percentage points.
However, it can reduce coverage and increase retrieval use. Overall accuracy improves by up to 15 percentage points on HotpotQA but falls by up to 17 percentage points on MuSiQue. These effects reflect a shift to a more selective, lower-risk operating point, not improved answers or confidence ranking.
A calibration map fitted before retrieval improves held-out calibration through retrieval depths one and two, but is worse than raw confidence at depth three for all three models. Additional evidence helps on average, but this aggregate effect does not establish whether confidence identifies which individual episodes will benefit from another retrieval.
Taken together, these results show that calibration can make commitment risk interpretable, but it does not estimate the expected benefit of another retrieval. Retrieval therefore requires a separate value-of-information or utility estimate. Evaluations should report held-out calibration, risk-coverage, and retrieval cost.
\end{abstract}

\maketitle

\section{Introduction}

\begin{figure*}[t]
  \centering
  \resizebox{\textwidth}{!}{%
  \begin{tikzpicture}[
    scale=1.14,
    transform shape,
    x=1cm,
    y=1cm,
    >=Latex,
    score/.style={
      circle,
      minimum size=0.68cm,
      inner sep=0pt,
      font=\sffamily\tiny\bfseries,
      line width=0.8pt
    },
    action/.style={
      rounded corners=3pt,
      minimum width=1.55cm,
      minimum height=0.58cm,
      align=center,
      font=\sffamily\tiny\bfseries,
      line width=0.8pt
    },
    note/.style={
      rounded corners=3pt,
      minimum width=2.14cm,
      minimum height=0.58cm,
      align=center,
      font=\sffamily\tiny,
      line width=0.7pt
    },
    lane/.style={
      rounded corners=5pt,
      minimum width=8.85cm,
      minimum height=0.88cm,
      line width=0.45pt
    },
    flow/.style={-{Latex[length=2.3mm,width=1.8mm]}, line width=1.15pt, draw=black}
  ]
    \coordinate (relorigin) at (-1.75,-0.92);
    \coordinate (reltop) at (-1.75,1.10);
    \coordinate (relright) at (0.10,-0.92);
    \draw[-{Latex[length=1.3mm,width=1.1mm]}, line width=0.5pt, draw=black]
      (relorigin) -- (reltop);
    \draw[-{Latex[length=1.3mm,width=1.1mm]}, line width=0.5pt, draw=black]
      (relorigin) -- (relright);
    \draw[dashed, line width=0.55pt, draw=black!35]
      (relorigin) -- (0.10,1.10);
    \draw[densely dotted, line width=0.55pt, draw=black!42]
      (-0.36,-0.92) -- (-0.36,0.09) -- (-1.75,0.09);
    \node[rectangle, rounded corners=0.4pt, minimum size=0.16cm, inner sep=0pt,
      draw=green!55!black, fill=green!55!black, line width=0.45pt]
      at (-1.75,0.09) {};
    \node[circle, minimum size=0.19cm, inner sep=0pt,
      draw=orange!90!black, fill=orange!45, line width=0.7pt]
      at (-0.36,0.09) {};
    \node[font=\sffamily\tiny\bfseries, text=black]
      at (-0.82,1.48) {RELIABILITY DIAGRAM};
    \node[font=\sffamily\tiny, text=black] at (-0.82,-1.30)
      {confidence};
    \node[font=\sffamily\tiny, text=black, rotate=90] at (-2.28,0.09)
      {accuracy};
    \node[font=\sffamily\tiny, text=orange!70!black, anchor=south]
      at (-0.36,0.35) {$(.75,\textcolor{green!40!black}{.50})$};
    \node[font=\sffamily\tiny, text=black, anchor=north east]
      at (-1.78,-0.94) {$0$};
    \node[font=\sffamily\tiny, text=black, anchor=east]
      at (-1.78,1.10) {$1$};
    \node[font=\sffamily\tiny, text=black, anchor=north]
      at (0.10,-0.94) {$1$};

    \draw[draw=black, line width=0.5pt] (1.22,1.20) -- (9.65,1.20);
    \node[font=\sffamily\tiny\bfseries, text=black]
      at (2.55,1.48) {SIGNAL};
    \node[font=\sffamily\tiny\bfseries, text=black]
      at (3.95,1.92) {FIXED POLICY [$\tau=0.7$]};
    \node[font=\sffamily\tiny\bfseries, text=black]
      at (5.65,1.48) {ACTION};
    \node[font=\sffamily\tiny\bfseries, text=black]
      at (8.28,1.48) {CONSEQUENCE};

    \node[font=\sffamily\tiny\bfseries, text=orange!75!black,
      anchor=west, align=left] at (0.86,0.55) {RAW\\SCORE};
    \node[score, minimum size=0.78cm, draw=orange!90!black, fill=orange!40,
      text=orange!70!black] (raw) at (2.55,0.55) {$0.75$};
    \node[font=\sffamily\tiny, text=black, align=center] (rawtest)
      at (3.95,0.55) {$0.75\geq\tau$};
    \node[action, draw=orange!90!black, fill=orange!38,
      text=orange!70!black] (commit) at (5.65,0.55) {COMMIT};
    \node[note, draw=red!58!black, fill=red!6, text=red!58!black]
      (risk) at (8.28,0.55) {empirical correctness\\$\approx 50\%$};
    \draw[flow] (raw) -- (rawtest);
    \draw[flow] (rawtest) -- (commit);
    \draw[flow] (commit) -- (risk);

    \node[font=\sffamily\tiny, text=green!45!black,
      anchor=west, align=left] at (0.42,-0.78) {\textbf{CALIBRATED}\\$p(\mathrm{correct})$};
    \node[score, minimum size=0.78cm, draw=green!70!black, fill=green!38,
      text=green!40!black] (cal) at (2.55,-0.78) {$0.50$};
    \node[font=\sffamily\tiny, text=black, align=center] (caltest)
      at (3.95,-0.78) {$0.50<\tau$};
    \node[action, draw=green!70!black, fill=green!35,
      text=green!40!black] (retrieve) at (5.65,-0.78) {RETRIEVE};
    \node[note, draw=black!42, fill=white, text=black]
      (tradeoff) at (8.28,-0.78) {cost $+1$\\value unknown};
    \draw[flow] (cal) -- (caltest);
    \draw[flow] (caltest) -- (retrieve);
    \draw[flow] (retrieve) -- (tradeoff);

    \node[rounded corners=2pt, draw=blue!55!black, fill=white,
      line width=0.6pt, inner xsep=4pt, inner ysep=1.5pt,
      font=\sffamily\tiny\bfseries, text=blue!55!black, anchor=west]
      at (2.99,-0.12) {CALIBRATE:\ $\phi(0.75)=\textcolor{green!40!black}{0.50}$};
    \draw[-{Latex[length=2.1mm,width=1.5mm]}, line width=0.9pt,
      draw=blue!55!black] (raw) -- (cal);
  \end{tikzpicture}%
  }
  \caption{In this example, a raw
  confidence of $0.75$ exceeds the commitment threshold even though comparable
  answers are correct only about half the time. Mapping the score to $0.50$
  triggers retrieval instead; the action has a cost, while its benefit is not
  implied by calibration.}
  \label{fig:motivating-example}
\end{figure*}
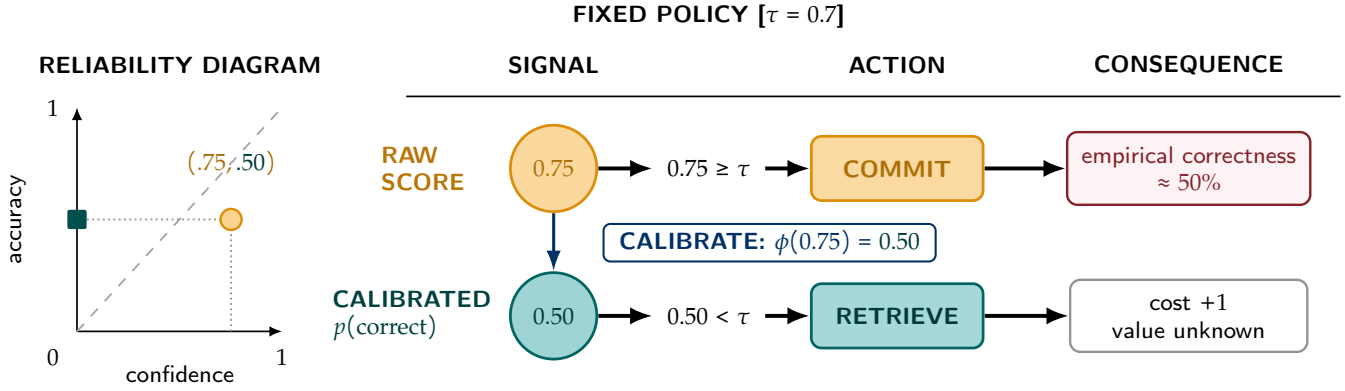

Interactive retrieval-augmented language-model agents must repeatedly decide whether their
current answer is reliable enough to return, whether another retrieval is worth
its cost, or whether to defer. Similar choices arise when agents query memory \citep{chhikara2025mem0}, use external knowledge \citep{chhikara2023knowledge}, or invoke specialized reasoning components \citep{allensound}. These systems often make the decision
by comparing model confidence with a fixed threshold. Consider an example demonstrated in Figure~\ref{fig:motivating-example}, where an agent reports
confidence $0.75$ and commits whenever confidence exceeds $0.7$. If answers near
$0.75$ are correct only half of the time, the threshold does not enforce the
intended risk requirement. Recalibration may move the score below $0.7$ and
trigger retrieval, but that action can consume resources without producing a
better answer. The central problem is therefore not confidence estimation alone,
but how the numerical meaning of confidence interacts with the policy that
consumes it.

This interaction separates \emph{monitoring} from \emph{control}. A confidence
judgment monitors the agent's current state of knowledge; commitment, retrieval,
and deferral are control actions based on that knowledge
\citep{nelson1990metamemory}. Calibration can make a score interpretable as an
empirical probability of current correctness, but it does not by itself specify
the value of another retrieval action. Nor does it improve how the score ranks
examples: a monotone calibration map preserves score order except when it
creates ties. A useful evaluation must therefore measure not only current-answer
calibration, but also the actions, outcomes, and costs that the signal induces.
Recent work studies calibration, selective prediction, and adaptive retrieval
largely as separate problems. We connect these problems by changing only the
score supplied to the controller. For each question, we first collect the
model's answer and confidence at each step of one predetermined evidence path.
We fit the calibration map on a separate split (calibration split), freeze it, and then replay the
saved states twice---once with raw confidence and once with calibrated
confidence. Because the answers, evidence, thresholds, budgets, and costs are
identical, any difference in actions comes only from the score shown to the
controller. Even though every depth is collected in advance, each replay sees
states only until its policy commits, abstains, or escalates; it never sees
future evidence after termination. This protocol is designed for controlled attribution rather than for reproducing the behavior of a production retriever. In real retrieval, relevant evidence can correct an answer, but distractors can waste budget or cause a correct answer to become wrong \citep{cuconasu2024power,liu2024lost}.

This design lets us test three linked questions: whether a fixed commitment
threshold behaves differently with calibrated confidence rather than raw confidence;
whether a calibration map fitted before retrieval remains reliable as evidence
changes the model's answer and confidence; and whether the resulting policy
changes improve accuracy among committed answers, coverage, overall accuracy,
and retrieval cost together. We do not collapse these outcomes into a single
notion of ``\textit{better control}'': without explicit costs for errors, retrieval, and
deferral, greater selectivity does not mean greater utility. Using the same
numerical threshold before and after calibration is intentional. This is not a
matched-risk or matched-coverage comparison; it checks how the same nominal
cutoff behaves when applied to an uncalibrated score versus an estimated
probability of correctness.

As a calibration case study, we evaluate \textittt{Mistral Small 4} (\mistrallogo),
\textittt{GPT-OSS-120B} (\openailogo), and \textittt{Qwen3-235B} (\qwenlogo)
on multi-hop question-answering datasets: HotpotQA \citep{yang2018hotpotqa} and MuSiQue \citep{trivedi2022musique}.
At the same numerical commitment threshold, calibration increases accuracy conditional on
commitment in all six model--dataset pairs, by up to 41 percentage points (pp).
However, its system-level effect reverses across datasets: overall accuracy
improves by up to 15pp on HotpotQA but falls by up to 17pp on
MuSiQue as coverage decreases and retrieval cost increases. A map fitted before
retrieval also improves held-out calibration through depths one and two, but is
worse than raw confidence at depth three for all the three models. This is an
evaluation warning: a signal validated on static, no-evidence states may cease
to be valid as an interactive trajectory evolves. Calibration can interpret
current-answer risk, but it supplies no explicit estimate of whether the next
observation is worth its cost.
Our contributions are threefold:
\begin{enumerate}
  \item We introduce matched trajectory replay, a controlled evaluation protocol
  that isolates the effect of a confidence-to-action mapping while holding
  candidate answer states, evidence points, budgets, and action costs fixed.
  \item We provide a calibration case study showing that a probabilistically
    meaningful commitment threshold can improve committed-answer accuracy while
    changing coverage, overall accuracy, and retrieval cost in non-uniform ways.
  \item We evaluate calibration across intermediate retrieval states and show
    why commitment-risk calibration should be paired with a separate,
    state-aware value-of-information or utility estimate for retrieval.
\end{enumerate}

\section{Related Work}
\label{sec:related-work}

\paragraph{Evaluation of interactive agents.}
Interactive agents should be evaluated not only by terminal answer quality, but
also by the intermediate states and actions that produce it, including
retrieval decisions, coverage, and cost. Our protocol contributes to this
trajectory-level view by comparing confidence-to-action mappings on matched
answer and evidence states.

\paragraph{Confidence estimation and calibration.}
Classical calibration methods, including temperature scaling and isotonic
regression, map scores to probabilities whose empirical correctness rates can
be inspected with reliability diagrams or calibration-error measures
\citep{guo2017calibration,naeini2015bayesian,zadrozny2002transforming}. For
language models, confidence can be elicited directly in natural language
\citep{lin2022uncertainty,tian2023calibration,mielke2022linguistic,xiong2024uncertainty, chhikara2025confidence},
estimated from sampling or semantic variation
\citep{wang2023selfconsistency,kuhn2023semantic,farquhar2024semantic}, or
obtained from self-evaluation signals such as P(True)
\citep{kadavath2022know}. Surveys organize these black-box and white-box
approaches \citep{geng2024survey,shorinwa2024survey}. We deliberately use a
simple verbalized score and a frozen post-hoc map: the question is not which
confidence estimator minimizes Expected Calibration Error (ECE), but whether recalibrating the same signal
changes a fixed controller's actions.

\paragraph{Selective prediction and metacognitive control.}
Selective prediction formalizes the risk--coverage trade-off by allowing a
model to accept or reject individual predictions
\citep{elyaniv2010foundations,geifman2017selective,geifman2019bias}; selective
question answering and refusal-aware language-model methods extend this idea
to open-ended answers \citep{kamath2020selective,zhang2024rtuning,wen2024limits}.
The metacognitive view similarly distinguishes monitoring knowledge from
controlling further search or response termination
\citep{nelson1990metamemory}. Our setting extends static accept-or-reject
decisions to a finite information-seeking loop. The connection is
computational, not psychological: we do not claim that the evaluated models
implement human metacognition, instead, we show that confidence acts as a monitoring signal
for retrieval and terminal actions.

\paragraph{Adaptive retrieval, routing, and Retrieval-Augmented Generation (RAG) reliability.}
Adaptive retrieval methods trigger search from token probabilities, learned
reflection signals, question complexity, or uncertainty-related features
\citep{jiang2023flare,asai2024selfrag,jeong2024adaptive,su2024dragin,
han2024uala,moskvoretskii2025adaptive}. Related routing systems allocate model
calls or tools to trade quality against computation
\citep{yao2023react,schick2023toolformer,chen2023frugalgpt,aggarwal2024automix,
ong2024routellm}. Calibration-oriented RAG changes the retrieved documents to
improve decision reliability \citep{jang2025calibrag}; this is complementary to
our intervention, which fixes candidate evidence in order to study when it is
requested. Retrieval quality is itself a confound: irrelevant passages can
degrade answers, long contexts are not used uniformly, and uncertainty can
shift after evidence is added \citep{cuconasu2024power,liu2024lost,
soudani2025uncertainty}. We address these effects by replaying the same
evidence trajectories and reporting answer, confidence, coverage, and cost
changes jointly.

\begin{figure}[t]
  \centering
  \resizebox{\linewidth}{!}{%
  \begin{tikzpicture}[
    >=Latex,
    flow/.style={-{Latex[length=1.7mm]}, line width=0.8pt, draw=black},
    shell/.style={
      rounded corners=2.5pt,
      draw=black,
      line width=0.75pt,
      fill=white
    },
    stage/.style={
      font=\scriptsize\bfseries,
      text=black,
      align=center
    },
    subtitle/.style={
      font=\scriptsize,
      text=black,
      align=center
    },
    chip/.style={
      rounded corners=1.6pt,
      draw=black,
      line width=0.5pt,
      font=\scriptsize\bfseries,
      inner xsep=4pt,
      inner ysep=2pt,
      align=center
    },
    replay/.style={
      rounded corners=2.5pt,
      draw=black,
      line width=0.7pt,
      text width=3.52cm,
      minimum height=0.78cm,
      align=center,
      inner sep=3.5pt
    },
    signal/.style={
      rounded corners=2pt,
      draw=black,
      line width=0.65pt,
      minimum height=0.64cm,
      font=\scriptsize,
      align=center,
      inner xsep=3.5pt,
      inner ysep=2.5pt
    },
    scorecard/.style={
      rounded corners=2.5pt,
      draw=black,
      line width=0.75pt,
      text width=2.70cm,
      minimum height=0.78cm,
      font=\footnotesize,
      align=center,
      inner sep=4pt
    },
    lane/.style={
      rounded corners=2pt,
      draw=black,
      line width=0.55pt,
      text width=3.64cm,
      minimum height=0.54cm,
      font=\scriptsize,
      align=center,
      inner sep=3pt
    }
  ]
    \node[
      shell,
      fill=cyan!5,
      minimum width=4.00cm,
      minimum height=2.65cm
    ] (collect) at (0,0) {};
    \node[
      shell,
      fill=black!2,
      minimum width=4.15cm,
      minimum height=2.65cm
    ] (replaypanel) at (4.78,0) {};
    \node[
      shell,
      fill=black!2,
      minimum width=3.60cm,
      minimum height=2.65cm
    ] (compare) at (9.48,0) {};

    \node[stage, text=cyan!45!black] at (0,0.98) {1.\enspace COLLECT ONCE};
    \node[subtitle] at (0,0.68) {$q$ + fixed evidence plan};
    \draw[flow] (-1.48,0.18) -- (-0.76,0.18);
    \draw[flow] (-0.43,0.18) -- (0.29,0.18);
    \draw[flow] (0.62,0.18) -- (1.34,0.18);
    \foreach \x/\state in {-1.64/$s_0$,-0.60/$s_1$,0.46/$s_2$,1.50/$s_3$} {
      \node[circle, draw=cyan!55!black, fill=cyan!16, line width=0.7pt,
        minimum size=0.32cm, inner sep=0pt, font=\scriptsize\bfseries,
        text=cyan!50!black] at (\x,0.18) {\state};
    }
    \node[font=\footnotesize, text=black] at (0,-0.34)
      {store $(\hat y_t, v_t, E_t, b_t)_{t=0}^{T}$};
    \node[font=\scriptsize\itshape, text=cyan!45!black] at (0,-0.74)
      {one trace per evidence plan};

    \node[stage] at (4.78,0.98) {2.\enspace REPLAY};
    \node[lane, fill=orange!32, text=orange!70!black]
      (rawscore) at (4.78,0.25) {\textbf{raw}\quad $v_t\ \longmapsto\ a_t^{\mathrm{raw}}$};
    \node[lane, fill=green!30, text=green!60!black]
      (calscore) at (4.78,-0.46) {\textbf{cal.}\quad $\kappa_t=\phi_M(v_t)\ \longmapsto\ a_t^{\mathrm{cal}}$};
    \node[font=\scriptsize\itshape, text=black] at (4.78,-0.96)
      {same stored states and policy $\pi$};

    \node[stage] at (9.48,0.98) {3.\enspace COMPARE TRACES};
    \node[chip, fill=orange!32, text=orange!70!black, minimum width=0.62cm]
      at (8.28,0.25) {raw};
    \draw[orange!70!black, line width=0.95pt]
      (8.72,0.25) -- (9.74,0.25);
    \node[font=\scriptsize, fill=black!2, inner xsep=3pt, text=orange!60!black]
      at (9.23,0.25) {$a^{\mathrm{raw}}_{0:T}$};
    \node[chip, fill=green!30, text=green!60!black, minimum width=0.62cm]
      at (8.28,-0.46) {cal.};
    \draw[green!45!black, line width=0.95pt]
      (8.72,-0.46) -- (9.74,-0.46);
    \node[font=\scriptsize, fill=black!2, inner xsep=3pt, text=green!45!black]
      at (9.23,-0.46) {$a^{\mathrm{cal}}_{0:T}$};
    \draw[-{Latex[length=1.2mm]}, black, line width=0.85pt]
      (9.74,0.25) -| (10.16,-0.08);
    \draw[-{Latex[length=1.2mm]}, black, line width=0.85pt]
      (9.74,-0.46) -| (10.16,-0.08);
    \node[chip, fill=black!3, text=black, minimum width=0.84cm,
      font=\scriptsize\bfseries] (delta) at (10.35,-0.08) {$\Delta a_{0:T}$};
    \draw[flow] ([xshift=0.18cm]delta.south) -- ++(0,-0.28);
    \node[font=\tiny, text=black, text width=1.65cm, align=center]
      at (10.68,-0.80) {\\coverage\\[-1pt] accuracy\\[-1pt] cost};

    \draw[flow] (collect.east) -- ++(0.28,0) |- (rawscore.west);
    \draw[flow] (collect.east) -- ++(0.28,0) |- (calscore.west);
    \draw[flow] (rawscore.east) -- ([xshift=0.24cm]compare.west |- rawscore.east);
    \draw[flow] (calscore.east) -- ([xshift=0.24cm]compare.west |- calscore.east);

  \end{tikzpicture}%
  }
  \caption{Matched trajectory replay. Candidate answer--confidence states are
  collected once for each fixed evidence plan and replayed through raw and calibrated
  policies.}
  \label{fig:controlled-replay}
\end{figure}
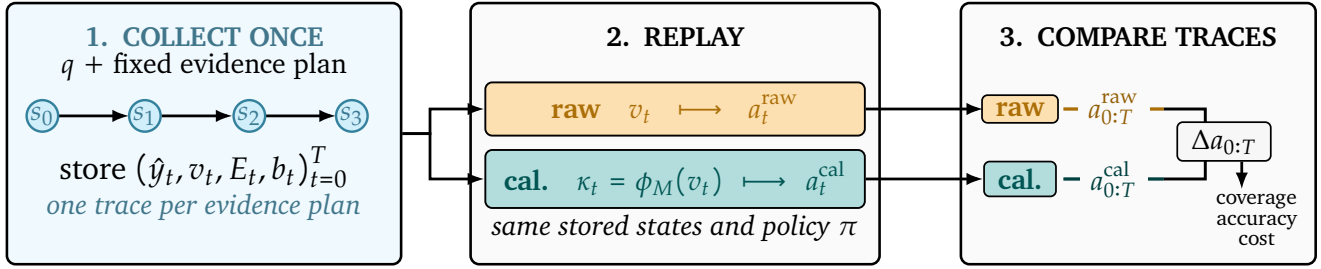

\section{Matched Trajectory Replay: Evaluation Protocol}
\label{sec:experiments}

Our question is narrow: \textit{\textbf{if two controllers see the same candidate
answers and retrieval opportunities, how do their actions change when they
receive different confidence scores?}} We answer it with matched trajectory
replay as shown in Figure~\ref{fig:controlled-replay}.
For every question, we first create one deterministic evidence plan and collect
the model's answer and confidence at each state on this plan. We fit the
calibration map only on the calibration split, freeze it, and then replay
the same stored final-test states through the two controllers. Therefore, any
difference between their traces is attributable to the score presented to the
controller, not to a different retriever call, generated answer, evidence
plan, or action cost. Collecting every state does not reveal future evidence to
a replay: each controller advances through the stored path only when it chooses
retrieval and terminates as soon as it commits, abstains, or escalates.

This section defines the fixed experimental objects: the evidence trajectories
and the collected traces. Section~\ref{sec:replay} then specifies how those
objects are replayed, which controllers are compared, and how their outcomes
are measured.

\subsection{Matched evidence trajectories}
\label{sec:trajectory-construction}

For question $q$, state $s_t=(q,E_t,b_t)$ contains cumulative evidence $E_t$
and remaining retrieval budget $b_t$. At each depth
$t\in\{0,1,2,3\}$, the model emits a short answer $\hat y_t$ and verbalized
confidence $v_t\in[0,1]$. Offline replay exposes every policy to the same answer
and evidence sequence at each possible retrieval depth. We use controlled
benchmark-grounded evidence plans rather than a live retriever. The primary policy and cost results use a fixed plan that combines dataset-marked supporting passages with ranked distractors which provides fixed evidence opportunities for paired replay and is intended for controlled attribution
rather than a simulation of a production retriever. The construction
below fixes what evidence is available after every possible retrieval action;
Appendix~\ref{app:elicitation} gives more details about the prompt used.

\paragraph{Constructing the fixed paths.}

For each question, we build one fixed retrieval plan from the dataset. The plan combines supporting passages that contain answer evidence with question-relevant distractors that appear useful but do not support the correct answer. We then divide these passages into three fixed retrieval slices, corresponding to increasing retrieval depth.
A retrieval action reveals the next complete slice, so the
visible evidence at depth $t$ is the cumulative first $t$ slices, with no
evidence at depth zero. We query the model at every depth before replay.
Consequently, either controller can change only the stored state at which it
terminates; it cannot change future evidence or a model response. The same
stored trajectories therefore support paired comparison in
Figure~\ref{fig:controlled-replay}. Appendix~\ref{app:evidence-paths}
specifies the support labels, distractor ranking, and slice allocation.

\subsection{Trace collection and labels}
\label{sec:trace-format}

To compare controllers on the same trajectories, we first collect every state
along each fixed evidence path. Each trace is associated with one dataset
example and one fixed evidence plan, and records the question, split, evidence
plan, depth-specific prompts and responses, verbalized confidence, and
correctness score. We generate traces deterministically using temperature zero
and three retrieval rounds. We assess answer correctness with the fixed
structured-output LLM judge \texttt{gemini-2.5-flash-lite}, which receives the
question, reference answers, and candidate answer. Appendix~\ref{app:judge-prompt}
provides the details about the judge prompt.

\section{Replay and Evaluation}
\label{sec:replay}

The replay stage operates exclusively on the traces defined in Section~\ref{sec:trace-format}, without invoking the generator, judge, or retriever. This setup enables paired comparisons of alternative confidence-to-action mappings over identical answer and evidence points.

\subsection{Counterfactual Confidence-to-Action Replay}

For the calibration, we fit a monotone isotonic map
$\phi_M$ on labeled depth-zero records from the calibration set, pooled
across the two datasets. We freeze this map
before final-test replay; no final-test trace contributes to the fit. The split
and collection settings are documented in Appendix~\ref{app:models-datasets}.

For every final-test trajectory, \rawmethod{} supplies the controller with the
stored raw confidence $v_t$, while \method{} supplies
$\kappa_t=\phi_M(v_t)$. Policies and threshold sweeps are then replayed
offline, so no evaluation call changes a recorded model response. Consequently,
Table~\ref{tab:model-results}, Figure~\ref{fig:risk-coverage}, and
Figure~\ref{fig:accuracy-cost} compare controllers on matched answer and
evidence states rather than independently collected trajectories. Because
isotonic regression preserves score order (apart from possible ties), it does
not improve ranking quality. However, it can rescale scores, so a fixed
threshold such as 0.7 selects a different set of states and thus results in a
different operating point.

\subsection{Controllers and baselines}

\paragraph{Controllers.} Both gated controllers use the same policy, thresholds
$(\tau_{\mathrm{lo}},\tau_{\mathrm{hi}})=(0.3,0.7)$, and maximum budget of
three retrieval actions. At any depth, the controller commits when confidence
is minimum $0.7$, else it retrieves while budget is not exhausted. Once the budget
is exhausted, it abstains at confidence $0.3$ or below and escalates when the scores
are between $0.3$ and $0.7$. To illustrate, the policy $\pi$ is defined as follows:
\begin{equation}
\pi(\kappa_t)=
\begin{cases}
\operatorname{commit}, & \kappa_t\geq\tau_{\mathrm{hi}},\\
\operatorname{retrieve}, & \kappa_t<\tau_{\mathrm{hi}}\ \text{and}\ b_t>0,\\
\operatorname{abstain}, & \kappa_t\leq\tau_{\mathrm{lo}}\ \text{and}\ b_t=0,\\
\operatorname{escalate}, & \text{otherwise.}
\end{cases}
\label{eq:policy}
\end{equation}
These thresholds were fixed a priori and not tuned on the test set:
$\tau_{\mathrm{hi}}=0.7$ represents a nominal 70\% correctness requirement for
commitment, while $\tau_{\mathrm{lo}}=0.3$ separates very low-confidence
exhausted-budget cases from escalation. For calibrated confidence, this
commitment threshold is intended to have a probabilistic interpretation; for raw
confidence, the identical number is only a heuristic. The fixed-number
comparison tests whether the commitment threshold retains the same practical
meaning; it does not compare controllers at matched risk or coverage. A
deployment threshold should instead be selected on a held-out validation set
under an explicit error--cost trade-off. Abstention and escalation are uncommitted terminal labels; we do not execute an
external escalation. Retrieval has unit cost, and commit, abstention, and
escalation have zero cost. 

\paragraph{Baselines.} Alongside the principal \rawmethod{}--\method{}
comparison, Parametric-only commits at depth zero without retrieval, while
Fixed RAG@$k$ reveals $k\in\{1,2,3\}$ evidence slices and then commits. The
Adaptive proxy is a deterministic query-complexity baseline: it chooses depth 3
for questions with at least 18 alphanumeric word tokens or at least two distinct
marker strings, depth 2 for at least nine such tokens or at least one marker
string, and depth 1 otherwise. The marker strings are \{same, both, whose,
which, who was, before, after\}; each marker type counts at most once.

\begin{table*}[t]
  \caption{Controlled-trajectory test results at the fixed operating point. All rows use
  fixed plans that interleave dataset-marked supporting passages with ranked
  distractors. OA$\uparrow$ is the percentage of all episodes ending in
  a correct commit; CA$\uparrow$ is accuracy conditional on commitment;
  Cov.$\uparrow$ is coverage; and Cost$\downarrow$ is the mean number of
  retrieval actions. Boldface marks
  the highest OA or CA within each model--dataset block. Blue highlight
  marks the higher OA or CA value between \method{} and \rawmethod{}. Fixed-depth and
  adaptive baselines always commit and therefore have 100\% coverage.}
  \label{tab:model-results}
  \centering
  \scriptsize
  \setlength{\tabcolsep}{3.4pt}
  \resizebox{\textwidth}{!}{%
  \begin{tabular}{llrrrrrrrr}
    \toprule
    & & \multicolumn{4}{c}{\textbf{HotpotQA} ($N=1{,}852$)}
        & \multicolumn{4}{c}{\textbf{MuSiQue} ($N=605$)} \\
    \cmidrule(lr){3-6}\cmidrule(lr){7-10}
    \textbf{Model} & \textbf{System}
      & \textbf{OA} & \textbf{CA} & \textbf{Cov.} & \textbf{Cost}
      & \textbf{OA} & \textbf{CA} & \textbf{Cov.} & \textbf{Cost} \\
    \midrule
    \multirow{7}{*}{\mistrallogo~Mistral Small 4 (2603)}
      & Parametric only & 39.1 & 39.1 & 100.0 & 0.00 & 10.6 & 10.6 & 100.0 & 0.00 \\
      & Fixed RAG@1    & 66.6 & 66.6 & 100.0 & 1.00 & 34.4 & 34.4 & 100.0 & 1.00 \\
      & Fixed RAG@2    & 66.4 & 66.4 & 100.0 & 2.00 & 41.2 & 41.2 & 100.0 & 2.00 \\
      & Fixed RAG@3    & \bestval{83.0} & 83.0 & 100.0 & 3.00 & \bestval{54.9} & \bestval{54.9} & 100.0 & 3.00 \\
      & Adaptive proxy & 72.4 & 72.4 & 100.0 & 2.32 & 52.4 & 52.4 & 100.0 & 2.88 \\
      & \rawmethod{}   & 42.1 & 42.1 & 100.0 & 0.07 &  \compareval{15.9}  & 15.9 & 99.7 & 0.26 \\
      & \method{}      & \compareval{46.3} & \compareval{\bestval{83.1}} & 55.7 & 2.44 & 3.3  & \compareval{45.5} & 7.3 & 2.92 \\
    \midrule
    \multirow{7}{*}{\openailogo~GPT-OSS-120B}
      & Parametric only & 51.6 & 51.6 & 100.0 & 0.00 & 21.0 & 21.0 & 100.0 & 0.00 \\
      & Fixed RAG@1    & 78.2 & 78.2 & 100.0 & 1.00 & 43.3 & 43.3 & 100.0 & 1.00 \\
      & Fixed RAG@2    & 77.1 & 77.1 & 100.0 & 2.00 & 48.1 & 48.1 & 100.0 & 2.00 \\
      & Fixed RAG@3    & \bestval{89.3} & \bestval{89.3} & 100.0 & 3.00 & \bestval{61.7} & 61.7 & 100.0 & 3.00 \\
      & Adaptive proxy & 82.1 & 82.1 & 100.0 & 2.32 & 59.2 & 59.2 & 100.0 & 2.88 \\
      & \rawmethod{}   & 70.7 & 70.9 & 99.6 & 0.44 & \compareval{45.6} & 49.9 & 91.4 & 1.16 \\
      & \method{}      & \compareval{78.7} & \compareval{89.0} & 88.5 & 1.43 & 28.9 & \compareval{\bestval{80.3}} & 36.0 & 2.59 \\
    \midrule
    \multirow{7}{*}{\qwenlogo~Qwen3-235B-A22B-2507}
      & Parametric only & 43.8 & 43.8 & 100.0 & 0.00 & 13.1 & 13.1 & 100.0 & 0.00 \\
      & Fixed RAG@1    & 71.1 & 71.1 & 100.0 & 1.00 & 35.2 & 35.2 & 100.0 & 1.00 \\
      & Fixed RAG@2    & 70.4 & 70.4 & 100.0 & 2.00 & 46.1 & 46.1 & 100.0 & 2.00 \\
      & Fixed RAG@3    & \bestval{87.5} & \bestval{87.5} & 100.0 & 3.00 & \bestval{60.3} & \bestval{60.3} & 100.0 & 3.00 \\
      & Adaptive proxy & 77.4 & 77.4 & 100.0 & 2.32 & 56.9 & 56.9 & 100.0 & 2.88 \\
      & \rawmethod{}   & 60.9 & 60.9 & 99.9 & 0.42 & \compareval{41.7} & 43.7 & 95.4 & 1.10 \\
      & \method{}      & \compareval{75.9} & \compareval{79.9} & 95.0 & 1.40 & 38.8 & \compareval{59.5} & 65.3 & 2.14 \\
    \bottomrule
  \end{tabular}
  }
\end{table*}

\section{Results}
\label{sec:results}

We ask five research questions (RQs). First, does calibration improve the
fixed-threshold controller, and what is the impact on coverage and retrieval?
Second, how do numerical thresholds change risk--coverage? Third, does the
calibration map transfer beyond its fit distribution? Fourth, do larger budgets
improve the accuracy--cost trade-off? Fifth, are the stored evidence transitions
actually useful?
Unless stated otherwise, policy and action results use the controlled,
benchmark-grounded test setup described in Section~\ref{sec:trajectory-construction}.
Held-out ECE in Table~\ref{tab:heldout-ece} pools all collected test states to
measure how the depth-zero calibration map transfers across retrieval depths.
HotpotQA and MuSiQue are reported separately in tables because their behavior
differs materially. For the figures, we combine all 1,852 HotpotQA and 605
MuSiQue test episodes into one set to show overall model-level trade-offs.

\paragraph{RQ1: What changes when the confidence-to-action mapping changes?}

We compare raw and calibrated confidence at the same numerical threshold,
$\tau_{\mathrm{hi}}=0.7$, on the test trajectories in Table~\ref{tab:model-results}. This isolates whether the threshold retains its
operational meaning after calibration. Exact checkpoint identifiers
and source-order split counts are given in Appendix~\ref{app:models-datasets}.

Table~\ref{tab:model-results} first shows the accuracy--coverage--cost trade-off
at this operating point.  The fixed-depth and adaptive baselines always commit,
so their overall accuracy (OA) equals their committed accuracy (CA) and their coverage is 100\%; Fixed RAG@3 attains the
highest OA in every model--dataset pair, but at a cost of three retrieval actions.
The gated controllers instead vary both coverage and retrieval cost to select a
subset of answers.  Accordingly, OA must be interpreted together with CA,
coverage, and cost: a higher CA need not yield a higher OA when substantially
fewer episodes are committed.

Against the \rawmethod{} controller, \method{} increases CA
in all six model--dataset pairs, by 15.8--41.0pp . This
selectivity comes with 4.9--92.4pp lower coverage and 0.98--2.66 additional
retrieval actions. The resulting OA change depends on the dataset. On HotpotQA,
it rises for all three models by 4.2--15.0pp; on MuSiQue, it
falls for all three by 2.9--16.7pp. For example, calibration raises
Mistral's MuSiQue CA from 15.9\% to 45.5\%, but reduces coverage from 99.7\% to
7.3\%; OA consequently falls from 15.9\% to 3.3\% while cost rises from 0.26 to
2.92 actions.

Thus, at a fixed numerical threshold, calibration consistently makes committed
answers more accurate, but it does not uniformly improve the end-to-end OA--cost
trade-off.  It changes the controller's operating point---in particular, its
commitment set, coverage, and retrieval use---rather than improving the
underlying answers or confidence ranking. This warns that committed-answer
accuracy alone can make a controller appear substantially better even when
coverage collapses, retrieval cost rises, and the end-to-end OA--cost outcome
degrades.

\begin{figure}[!ht]
  \centering
  \begin{minipage}[t]{0.355\linewidth}
  \centering
  \begin{tikzpicture}
    \begin{axis}[
      width=\linewidth,
      height=0.88\linewidth,
      xmin=0,
      xmax=1.02,
      ymin=0,
      ymax=0.80,
      xlabel={Coverage},
      ylabel={Selective risk},
      grid=both,
      major grid style={draw=black!12},
      minor grid style={draw=black!6},
      minor tick num=1,
      tick label style={font=\footnotesize},
      label style={font=\footnotesize},
      axis line style={draw=black!55},
      tick style={draw=black!55},
      clip=false
    ]
      \node[anchor=north west, inner sep=0pt] at (rel axis cs:0.02,0.98)
        {\mistralfiglogo};
      \addplot+[raw-gated curve] coordinates {
        (1.000000,0.660155)
        (0.999593,0.655537)
        (0.999593,0.655537)
        (0.999186,0.643177)
        (0.980057,0.479651)
        (0.973545,0.455686)
        (0.437525,0.184186)
      };
      \addplot+[calibrated-gated curve] coordinates {
        (0.980057,0.479651)
        (0.973545,0.455686)
        (0.437525,0.184186)
        (0.437525,0.184186)
      };
      \addplot+[baseline markers] coordinates {
        (1.000000,0.678877)
        (1.000000,0.413512)
        (1.000000,0.398453)
        (1.000000,0.239316)
        (1.000000,0.325193)
      };
      \addplot+[raw operating point] coordinates {(0.999186,0.643177)};
      \addplot+[calibrated operating point] coordinates {(0.437525,0.184186)};
      \node[
        font=\tiny,
        text=orange!70!black,
        anchor=south east
      ] at (axis cs:0.985,0.675) {raw $\tau=0.7$};
      \node[
        font=\tiny,
        text=green!45!black,
        anchor=south west,
        align=left
      ] at (axis cs:0.450,0.215) {cal $\tau=0.7$};
    \end{axis}
  \end{tikzpicture}
  \end{minipage}\subplotgap%
  \begin{minipage}[t]{0.355\linewidth}
  \centering
  \begin{tikzpicture}
    \begin{axis}[
      width=\linewidth,
      height=0.88\linewidth,
      xmin=0,
      xmax=1.02,
      ymin=0,
      ymax=0.80,
      xlabel={Coverage},
      grid=both,
      major grid style={draw=black!12},
      minor grid style={draw=black!6},
      minor tick num=1,
      tick label style={font=\footnotesize},
      label style={font=\footnotesize},
      clip=false,
      axis line style={draw=black!55},
      tick style={draw=black!55}
    ]
      \node[anchor=north west, inner sep=0pt] at (rel axis cs:0.02,0.98)
        {\openaifiglogo};
      \addplot+[raw-gated curve] coordinates {
        (1.000000,0.555149)
        (1.000000,0.517298)
        (1.000000,0.481888)
        (0.994302,0.403193)
        (0.877493,0.204082)
        (0.749288,0.118957)
        (0.249898,0.050489)
      };

      \addplot+[calibrated-gated curve] coordinates {
        (0.975987,0.339033)
        (0.962556,0.307400)
        (0.853887,0.176835)
        (0.755800,0.120625)
        (0.749288,0.118957)
        (0.260480,0.050000)
      };

      \addplot+[baseline markers] coordinates {
        (1.000000,0.559219)
        (1.000000,0.303622)
        (1.000000,0.300366)
        (1.000000,0.175417)
        (1.000000,0.235246)
      };
      \addplot+[raw operating point] coordinates {(0.975987,0.339033)};
      \addplot+[calibrated operating point] coordinates {(0.755800,0.120625)};
      \node[
        font=\tiny,
        text=orange!70!black,
        anchor=south east
      ] at (axis cs:0.985,0.375) {raw $\tau=0.7$};
      \node[
        font=\tiny,
        text=green!45!black,
        anchor=south east,
        align=right
      ] at (axis cs:0.770,0.155) {cal $\tau=0.7$};
  \end{axis}
  \end{tikzpicture}
  \end{minipage}\subplotgap%
  \begin{minipage}[t]{0.355\linewidth}
  \centering
  \begin{tikzpicture}
    \begin{axis}[
      width=\linewidth,
      height=0.88\linewidth,
      xmin=0,
      xmax=1.02,
      ymin=0,
      ymax=0.80,
      xlabel={Coverage},
      grid=both,
      major grid style={draw=black!12},
      minor grid style={draw=black!6},
      minor tick num=1,
      tick label style={font=\footnotesize},
      label style={font=\footnotesize},
      clip=false,
      axis line style={draw=black!55},
      tick style={draw=black!55}
    ]
      \node[anchor=north west, inner sep=0pt] at (rel axis cs:0.02,0.98)
        {\qwenfiglogo};
      \addplot+[raw-gated curve] coordinates {
        (1.000000,0.596256)
        (0.997151,0.567347)
        (0.997151,0.565306)
        (0.987790,0.431809)
        (0.894180,0.277651)
        (0.877086,0.238515)
        (0.824990,0.221510)
      };
      \addplot+[calibrated-gated curve] coordinates {
        (0.989011,0.446091)
        (0.987383,0.428689)
        (0.894180,0.277651)
        (0.877086,0.238515)
      };
      \addplot+[baseline markers] coordinates {
        (1.000000,0.637770)
        (1.000000,0.377289)
        (1.000000,0.355718)
        (1.000000,0.191697)
        (1.000000,0.276760)
      };
      \addplot+[raw operating point] coordinates {(0.987790,0.431809)};
      \addplot+[calibrated operating point] coordinates {(0.877086,0.238515)};
      \node[
        font=\tiny,
        text=orange!70!black,
        anchor=south east
      ] at (axis cs:0.975,0.470) {raw $\tau=0.7$};
      \node[
        font=\tiny,
        text=green!45!black,
        anchor=south east,
        align=right
      ] at (axis cs:0.900,0.175) {cal $\tau=0.7$};
    \end{axis}
  \end{tikzpicture}
  \end{minipage}
  \par\vspace{-0.25em}
  \figurelegend
  \caption{Fixed numerical thresholds on the risk--coverage plane for the
  micro-pooled controlled-trajectory tests ($N=2{,}457$:
  1{,}852 HotpotQA and 605
  MuSiQue). Raw and calibrated lines apply the same set of up to seven numeric
  thresholds; they are sampled operating points, not full or matched-coverage
  frontiers. Circles mark $\tau=0.7$ and squares show confidence-blind baselines.}
  \label{fig:risk-coverage}
\end{figure}
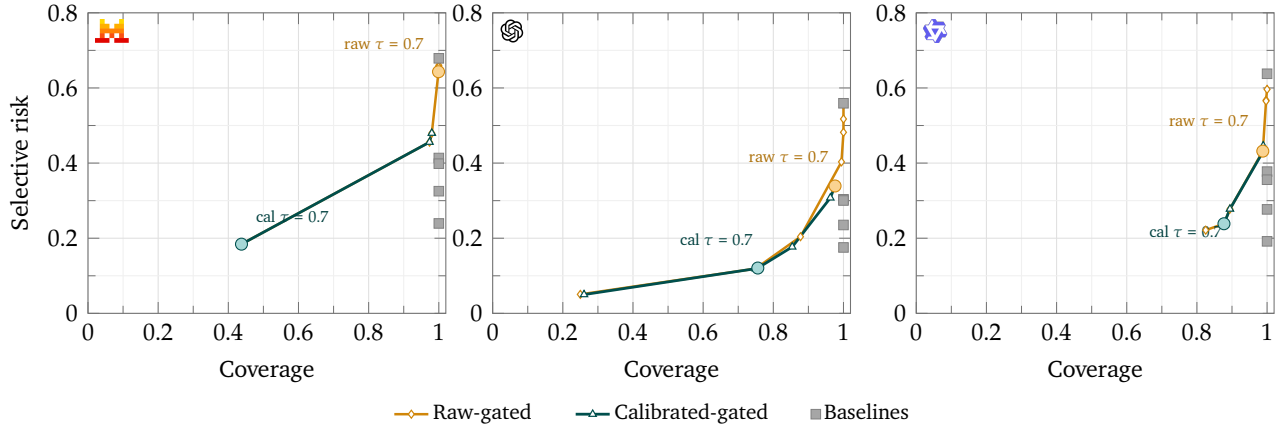

\paragraph{RQ2: How does calibration move the risk--coverage operating point?}

Figure~\ref{fig:risk-coverage} directly compares \rawmethod{} and \method{}
at shared numerical thresholds of 0.20, 0.33, 0.50, 0.67, 0.80, 0.90, and
0.95. The fixed $\tau=0.7$ operating point is marked separately. For both controllers,
increasing the threshold generally reduces coverage and selective risk
($1-\mathrm{CA}$). However, applying the same numeric threshold to raw and
calibrated confidence selects different operating points.

On the micro-pooled test set, \method{} generally moves to a more selective
operating point than \rawmethod{}: it commits fewer answers, but those answers
have lower selective risk. This shift is largest for Mistral, intermediate for
GPT-OSS, and smallest for Qwen. At $\tau=0.7$, Mistral moves from 35.7\% CA /
99.9\% coverage to 81.6\% CA / 43.8\% coverage; GPT-OSS moves from 66.1\% CA /
97.6\% coverage to 87.9\% CA / 75.6\% coverage; and Qwen moves from 56.8\% CA /
98.8\% coverage to 76.1\% CA / 87.7\% coverage.
The squares provide a full-coverage reference: Fixed RAG@3 is the lowest-risk
confidence-blind baseline in all three panels. They underscore that a gated
point's selective risk must be interpreted together with its coverage.

Because isotonic regression preserves score order (apart from possible ties),
it does not improve ranking quality. In our experiments, it rescales confidence
so that the same numerical threshold yields a more selective commitment set,
with lower observed risk and lower coverage. Interactive-agent evaluations
should therefore report risk--coverage operating points rather than selective
accuracy alone: lower risk at a fixed threshold can reflect a narrower
commitment set rather than an improved underlying ranking.

\begin{table*}[h]
  \caption{Held-out 10-bin ECE (percentage points) on test states.
  Depth zero is invariant to retrieved evidence; later depths pool all
  collected evidence paths and micro-average HotpotQA and MuSiQue. Calibration improves
  ECE through depth two, but at
  depth three calibrated ECE exceeds raw ECE for all three models.}
  \label{tab:heldout-ece}
  \centering
  \small
  \setlength{\tabcolsep}{5pt}
  \begin{tabular}{lrrrrrrrr}
    \toprule
    & \multicolumn{2}{c}{\textbf{Depth 0}} & \multicolumn{2}{c}{\textbf{Depth 1}}
    & \multicolumn{2}{c}{\textbf{Depth 2}} & \multicolumn{2}{c}{\textbf{Depth 3}} \\
    \cmidrule(lr){2-3}\cmidrule(lr){4-5}\cmidrule(lr){6-7}\cmidrule(lr){8-9}
    \textbf{Model} & \textbf{Raw} & \textbf{Cal.} & \textbf{Raw} & \textbf{Cal.} & \textbf{Raw} & \textbf{Cal.} & \textbf{Raw} & \textbf{Cal.} \\
    \midrule
    \mistrallogo~Mistral Small 4 (2603) & 44.2 & 2.5 & 27.4 & 18.1 & 27.3 & 16.7 & 15.4 & 23.0 \\
    \openailogo~GPT-OSS-120B            & 27.7 & 2.5 & 12.2 & 9.5  & 12.7 & 8.6  & 6.1 & 8.0 \\
    \qwenlogo~Qwen3-235B-A22B-2507      & 25.5 & 2.0 & 22.0 & 5.5  & 21.0 & 7.0  & 12.8 & 14.6 \\
    \bottomrule
  \end{tabular}
\end{table*}

\paragraph{RQ3: Does a static calibration map remain valid as an interactive trajectory evolves?}

ECE summarizes the frequency-weighted gap between
average confidence and accuracy in ten equal-width confidence bins.
Table~\ref{tab:heldout-ece} reports this held-out metric on test states by
evidence depth. At depth zero, isotonic calibration reduces ECE from
25.5--44.2pp to 2.0--2.5pp. On the held-out trajectories, the frozen map
continues to improve ECE through depths one and two, with reductions of
2.7--16.5pp. At depth three, calibrated ECE is higher than raw ECE for all
three models, by 1.8--7.6pp. Thus, within these three-step benchmark-grounded
trajectories, a map fitted on static, depth-zero states improves calibration
through the first two retrieval steps but no longer does so at the third. This
is an evaluation warning: validating a signal on static states does not
establish its validity throughout an evolving interactive trajectory.

The depth-three reversal does not mean that retrieval necessarily makes the
answers worse. Rather, it suggests that retrieved evidence changes the
confidence--correctness relationship while the controller continues to use a
map fitted before any evidence was shown. A single depth-independent map can
therefore overcorrect high-confidence retrieved states.

Because Table~\ref{tab:heldout-ece} pools all stored test states across HotpotQA and MuSiQue,
it characterizes state-level calibration rather than the calibration of a policy-induced commitment set.
It therefore neither implies 70\% committed accuracy at a threshold of 0.7 for either dataset nor
rules out dataset-specific or policy-conditional deviations.

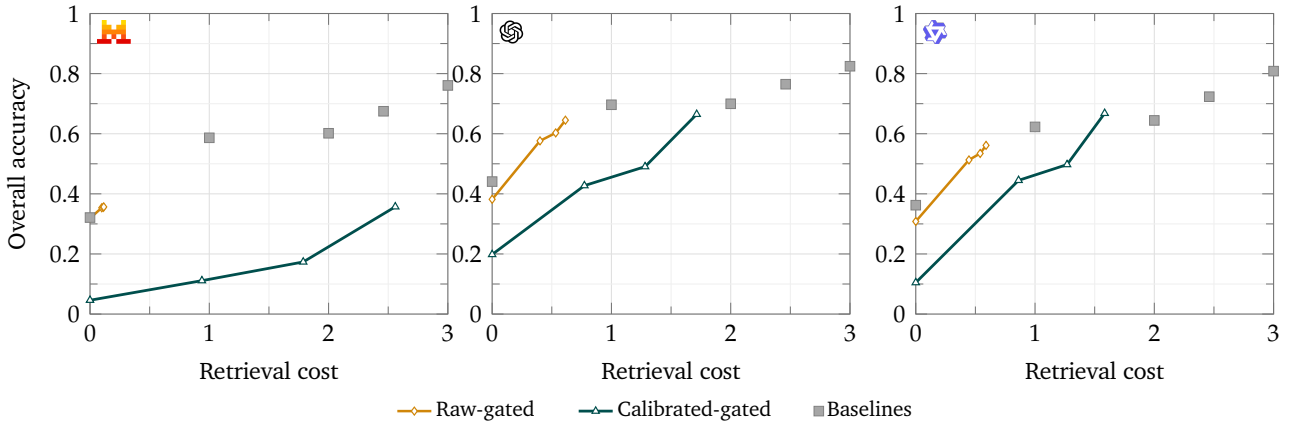
\begin{figure}[!ht]
  \centering
  \begin{minipage}[t]{0.355\linewidth}
  \centering
  \begin{tikzpicture}
    \begin{axis}[
      width=\linewidth,
      height=0.88\linewidth,
      xmin=0,
      xmax=3,
      ymin=0,
      ymax=1.00,
      xlabel={Retrieval cost},
      ylabel={Overall accuracy},
      grid=both,
      major grid style={draw=black!12},
      minor grid style={draw=black!6},
      minor tick num=1,
      tick label style={font=\footnotesize},
      label style={font=\footnotesize},
      axis line style={draw=black!55},
      tick style={draw=black!55}
    ]
      \node[anchor=north west, inner sep=0pt] at (rel axis cs:0.02,0.98)
        {\mistralfiglogo};
      \addplot+[raw-gated curve] coordinates {
        (0.000000,0.318681)
        (0.098494,0.353683)
        (0.110704,0.354904)
        (0.115181,0.356532)
      };
      \addplot+[calibrated-gated curve] coordinates {
        (0.000000,0.045991)
        (0.938136,0.111111)
        (1.787546,0.173382)
        (2.559219,0.356939)
      };
      \addplot+[baseline markers] coordinates {
        (0.000000,0.321123)
        (1.000000,0.586488)
        (2.000000,0.601547)
        (3.000000,0.760684)
        (2.460724,0.674807)
      };
    \end{axis}
  \end{tikzpicture}
  \end{minipage}\subplotgap%
  \begin{minipage}[t]{0.355\linewidth}
  \centering
  \begin{tikzpicture}
    \begin{axis}[
      width=\linewidth,
      height=0.88\linewidth,
      xmin=0,
      xmax=3,
      ymin=0,
      ymax=1.00,
      xlabel={Retrieval cost},
      grid=both,
      major grid style={draw=black!12},
      minor grid style={draw=black!6},
      minor tick num=1,
      tick label style={font=\footnotesize},
      label style={font=\footnotesize},
      clip=false,
      axis line style={draw=black!55},
      tick style={draw=black!55}
    ]
      \node[anchor=north west, inner sep=0pt] at (rel axis cs:0.02,0.98)
        {\openaifiglogo};
      \addplot+[raw-gated curve] coordinates {
        (0.000000,0.381766)
        (0.401302,0.576313)
        (0.532764,0.602768)
        (0.614164,0.645096)
      };

      \addplot+[calibrated-gated curve] coordinates {
        (0.000000,0.198209)
        (0.773301,0.427350)
        (1.282051,0.490435)
        (1.714693,0.664632)
      };

      \addplot+[baseline markers] coordinates {
        (0.000000,0.440781)
        (1.000000,0.696378)
        (2.000000,0.699634)
        (3.000000,0.824583)
        (2.460724,0.764754)
      };
  \end{axis}
  \end{tikzpicture}
  \end{minipage}\subplotgap%
  \begin{minipage}[t]{0.355\linewidth}
  \centering
  \begin{tikzpicture}
    \begin{axis}[
      width=\linewidth,
      height=0.88\linewidth,
      xmin=0,
      xmax=3,
      ymin=0,
      ymax=1.00,
      xlabel={Retrieval cost},
      grid=both,
      major grid style={draw=black!12},
      minor grid style={draw=black!6},
      minor tick num=1,
      tick label style={font=\footnotesize},
      label style={font=\footnotesize},
      clip=false,
      axis line style={draw=black!55},
      tick style={draw=black!55}
    ]
      \node[anchor=north west, inner sep=0pt] at (rel axis cs:0.02,0.98)
        {\qwenfiglogo};
      \addplot+[raw-gated curve] coordinates {
        (0.000000,0.308099)
        (0.446072,0.512821)
        (0.540090,0.534392)
        (0.589337,0.561254)
      };
      \addplot+[calibrated-gated curve] coordinates {
        (0.000000,0.104599)
        (0.862027,0.444851)
        (1.268620,0.497354)
        (1.583639,0.667888)
      };
      \addplot+[baseline markers] coordinates {
        (0.000000,0.362230)
        (1.000000,0.622711)
        (2.000000,0.644282)
        (3.000000,0.808303)
        (2.460724,0.723240)
      };
    \end{axis}
  \end{tikzpicture}
  \end{minipage}
  \par\vspace{-0.25em}
  \figurelegend
  \caption{OA--cost trade-off on the micro-pooled controlled-trajectory test
  set. Curves
  vary the maximum retrieval budget from 0--3; realized cost can be lower when
  the controller commits early. Squares denote no retrieval, Fixed RAG@1/@2,
  the adaptive proxy, and Fixed RAG@3.}
  \label{fig:accuracy-cost}
\end{figure}

\paragraph{RQ4: Does calibration improve the accuracy--cost trade-off?}

Figure~\ref{fig:accuracy-cost} directly compares \rawmethod{} and \method{}
as the maximum retrieval budget increases from zero to three. Realized cost can
be lower than the budget when a controller commits early, so each point is read
as an OA--cost operating point rather than as a fixed-depth result. At budget
three, calibration leaves Mistral's OA unchanged at 35.7\% while raising mean
retrieval cost from 0.12 to 2.56 actions per episode. For GPT-OSS, it gains 2.0
OA points at 1.10 additional actions; for Qwen, it gains 10.7 points at 0.99
additional actions.

The squares provide the confidence-blind context: Fixed RAG@3 reaches
76.1--82.5\% OA at cost 3, and
the adaptive proxy reaches 67.5--76.5\% at cost 2.46. The calibrated Mistral
and GPT-OSS endpoints have lower OA and higher cost than Fixed RAG@1. Qwen's
endpoint exceeds Fixed RAG@1 by 4.5 percentage points of OA at 0.58 additiosnal cost; compared
with Fixed RAG@2, it has 2.4 percentage points higher OA at 0.42 lower cost.

The steeper calibrated sweeps do not by themselves indicate greater retrieval
efficiency: calibration begins from a substantially more selective zero-budget
policy and uses more retrieval actions as the budget grows. Overall,
calibration does not uniformly improve the OA--cost trade-off. It leaves
Mistral's OA unchanged at much higher cost, results in only a small gain for
GPT-OSS, and provides a substantial gain only for Qwen; even there, it does not
surpass the adaptive proxy.

\begin{table}[!htbp]
  \caption{Adjacent retrieval-transition diagnostics on micro-pooled
  controlled-trajectory test traces. Helpful denotes
  wrong$\rightarrow$correct, harmful
  denotes correct$\rightarrow$wrong, and Net (pp) is the helpful-minus-harmful
  percentage-point difference. Each model has 7{,}367 retrieval transitions
  that reveal new evidence; rates are percentages of transitions.}
  \label{tab:retrieval-transitions}
  \centering
  \small
  \setlength{\tabcolsep}{7pt}
  \begin{tabular}{lrrr}
    \toprule
    \textbf{Model} & \textbf{Helpful} & \textbf{Harmful} & \textbf{Net (pp)} \\
    \midrule
    \mistrallogo~Mistral Small 4 (2603) & 18.4 & 3.7 & $+14.7$ \\
    \openailogo~GPT-OSS-120B            & 16.2 & 3.4 & $+12.8$ \\
    \qwenlogo~Qwen3-235B-A22B-2507      & 18.4 & 3.6 & $+14.9$ \\
    \midrule
    Average                             & 17.7 & 3.6 & $+14.1$ \\
    \bottomrule
  \end{tabular}
\end{table}

\paragraph{RQ5: Is another evidence step useful?}

Across all stored adjacent state pairs with a nonempty next evidence slice,
regardless of replay policy, Table~\ref{tab:retrieval-transitions} compares
correctness before and after one additional evidence step. Averaged across
models, 17.7\% of transitions change an incorrect answer to a correct one,
whereas 3.6\% change a correct answer to an incorrect one. Helpful
transitions therefore substantially outnumber harmful ones, yielding an average
net shift of +14.1 percentage points.

These aggregate gains help explain why Fixed RAG@3 performs well: in these
adjacent test transitions, additional evidence often improves the model's
answer. However, this does not mean that confidence alone can decide when to
retrieve. Confidence at a given state is intended to estimate whether the
\emph{current} answer is correct, whereas retrieval control requires estimating
whether the \emph{next} evidence step is likely to help. A low-confidence answer
may receive little useful new information, and a moderately confident answer may
still benefit from multi-hop evidence. Calibration can make current-answer risk
easier to interpret, but it does not turn confidence into a
value-of-information estimate. Average retrieval helpfulness therefore does not
show that a confidence signal can identify which individual episodes should
retrieve next; evaluation must distinguish evidence quality from value-of-
information prediction.

\section{Discussion and Conclusion}
\label{sec:discussion}

Matched trajectory replay shows that calibration changes the decisions made by
a fixed threshold, not the stored answers. It improves committed-answer
accuracy by moving the controller to a more selective operating point, but its
effects on overall accuracy and retrieval cost vary by dataset and model. The
depth-zero map improves held-out calibration through the first two retrieval
depths but overcorrects high-confidence states at depth three. Although another
evidence step helps on average, this aggregate analysis does not establish that
confidence identifies which individual episodes will benefit.

The resulting design principle is to use calibration to make commitment risk
interpretable, not as a complete retrieval controller. Interactive-agent
evaluations should therefore report calibration and action outcomes at
intermediate trajectory states, not only at the initial state or terminal
outcome. Deployment should
validate calibration and select thresholds for each target domain under an
explicit error--cost objective, while a separate estimate of expected
information gain or utility governs retrieval. Our controlled replay isolates
the effect of the calibration map under a fixed policy; it does not
establish that a fixed $0.7$ threshold, or confidence gating alone, is optimal
for a live agent.

\section{Limitations}
\label{sec:limitations}

Controlled replay strengthens attribution but limits external validity. We use deterministic, benchmark-grounded evidence paths rather than a live production retriever, and we fix future evidence and queries. The replay therefore isolates the effect of replacing raw confidence with a fixed calibration map, but cannot capture how a different action would change the next query, retrieved passages, or model response in a live agent. Because the evidence construction uses benchmark support and ranked distractors, it can overstate retriever reliability.

The empirical scope is limited to two multi-hop QA datasets, three models, one deterministic response collection per model, and deterministic source-order splits. The method also assumes that a calibration map fitted at depth zero remains valid after evidence acquisition, an assumption that fails at depth three in our trajectories. Deployment systems would need depth- or state-conditioned calibration, or periodic recalibration using newly labeled later-state data.

Finally, correctness uses an LLM judge without a human-validation study. We treat abstention and unexecuted escalation as zero-credit outcomes, assign them zero cost, and price only retrieval actions, so the reported OA--cost curves are not end-to-end utility curves. Deployment should validate the judge, price every terminal action, and monitor calibration, coverage, and subgroup effects under shift.

\bibliography{main}

@inproceedings{allensound,
  title={Sound and Complete Neurosymbolic Reasoning with LLM-Grounded Interpretations},
  author={Allen, Bradley P and Chhikara, Prateek and Ferguson, Thomas Macaulay and Ilievski, Filip and Groth, Paul},
  booktitle={19th International Conference on Neurosymbolic Learning and Reasoning},
  year={2025},
}

@article{aggarwal2024automix,
  title={Automix: Automatically mixing language models},
  author={Aggarwal, Pranjal and Madaan, Aman and Anand, Ankit and Potharaju, Srividya Pranavi and Mishra, Swaroop and Zhou, Pei and Gupta, Aditya and Rajagopal, Dheeraj and Kappaganthu, Karthik and Yang, Yiming and others},
  journal={arXiv preprint arXiv:2310.12963},
  year={2023}
}

@inproceedings{asai2024selfrag,
  title={Self-rag: Learning to retrieve, generate, and critique through self-reflection},
  author={Asai, Akari and Wu, Zeqiu and Wang, Yizhong and Sil, Avi and Hajishirzi, Hannaneh},
  booktitle={International conference on learning representations},
  volume={2024},
  pages={9112--9141},
  year={2024}
}

@article{chen2023frugalgpt,
  title={Frugalgpt: How to use large language models while reducing cost and improving performance},
  author={Chen, Lingjiao and Zaharia, Matei and Zou, James},
  journal={arXiv preprint arXiv:2305.05176},
  year={2023}
}

@article{chhikara2025confidence,
  title={Mind the Confidence Gap: Overconfidence, Calibration, and Distractor Effects in Large Language Models},
  author={Chhikara, Prateek},
  journal={Transactions on Machine Learning Research},
  year={2025},
  doi={10.48550/arXiv.2502.11028},
  url={https://arxiv.org/abs/2502.11028}
}

@inproceedings{chhikara2023knowledge,
  title={Knowledge-Enhanced Agents for Interactive Text Games},
  author={Chhikara, Prateek and Zhang, Jiarui and Ilievski, Filip and Francis, Jonathan and Ma, Kaixin},
  booktitle={Proceedings of the 12th Knowledge Capture Conference 2023},
  series={K-CAP '23},
  pages={157--165},
  year={2023},
  publisher={Association for Computing Machinery},
  address={New York, NY, USA},
  doi={10.1145/3587259.3627561},
  url={https://doi.org/10.1145/3587259.3627561}
}

@inproceedings{chhikara2025mem0,
  title={Mem0: Building Production-Ready AI Agents with Scalable Long-Term Memory},
  author={Chhikara, Prateek and Khant, Dev and Aryan, Saket and Singh, Taranjeet and Yadav, Deshraj},
  booktitle={ECAI 2025},
  series={Frontiers in Artificial Intelligence and Applications},
  volume={413},
  pages={2993--3000},
  year={2025},
  doi={10.3233/FAIA251160},
  url={https://ebooks.iospress.nl/doi/10.3233/FAIA251160}
}

@inproceedings{cuconasu2024power,
  title={The power of noise: Redefining retrieval for rag systems},
  author={Cuconasu, Florin and Trappolini, Giovanni and Siciliano, Federico and Filice, Simone and Campagnano, Cesare and Maarek, Yoelle and Tonellotto, Nicola and Silvestri, Fabrizio},
  booktitle={Proceedings of the 47th International ACM SIGIR Conference on Research and Development in Information Retrieval},
  pages={719--729},
  year={2024}
}

@article{elyaniv2010foundations,
  title={On the Foundations of Noise-free Selective Classification.},
  author={El-Yaniv, Ran and others},
  journal={Journal of Machine Learning Research},
  volume={11},
  number={5},
  year={2010}
}

@article{farquhar2024semantic,
  title={Detecting hallucinations in large language models using semantic entropy},
  author={Farquhar, Sebastian and Kossen, Jannik and Kuhn, Lorenz and Gal, Yarin},
  journal={Nature},
  volume={630},
  number={8017},
  pages={625--630},
  year={2024},
  publisher={Nature Publishing Group UK London}
}

@article{geifman2017selective,
  title={Selective classification for deep neural networks},
  author={Geifman, Yonatan and El-Yaniv, Ran},
  journal={Advances in neural information processing systems},
  volume={30},
  year={2017}
}

@inproceedings{geifman2019bias,
  title={Bias-Reduced Uncertainty Estimation for Deep Neural Classifiers},
  author={Geifman, Yonatan and Uziel, Guy and El-Yaniv, Ran},
  booktitle={International Conference on Learning Representations},
  note = {ICLR}
}

@inproceedings{geng2024survey,
  title={A survey of confidence estimation and calibration in large language models},
  author={Geng, Jiahui and Cai, Fengyu and Wang, Yuxia and Koeppl, Heinz and Nakov, Preslav and Gurevych, Iryna},
  booktitle={Proceedings of the 2024 Conference of the North American Chapter of the Association for Computational Linguistics: Human Language Technologies (Volume 1: Long Papers)},
  pages={6577--6595},
  year={2024}
}

@inproceedings{guo2017calibration,
  title={On calibration of modern neural networks},
  author={Guo, Chuan and Pleiss, Geoff and Sun, Yu and Weinberger, Kilian Q},
  booktitle={International conference on machine learning},
  pages={1321--1330},
  year={2017},
  organization={PMLR}
}

@inproceedings{han2024uala,
  title={Towards uncertainty-aware language agent},
  author={Han, Jiuzhou and Buntine, Wray and Shareghi, Ehsan},
  booktitle={Findings of the Association for Computational Linguistics: ACL 2024},
  pages={6662--6685},
  year={2024}
}

@inproceedings{jeong2024adaptive,
  title={Adaptive-rag: Learning to adapt retrieval-augmented large language models through question complexity},
  author={Jeong, Soyeong and Baek, Jinheon and Cho, Sukmin and Hwang, Sung Ju and Park, Jong C},
  booktitle={Proceedings of the 2024 Conference of the North American Chapter of the Association for Computational Linguistics: Human Language Technologies (Volume 1: Long Papers)},
  pages={7036--7050},
  year={2024}
}

@inproceedings{jiang2023flare,
  title={Active retrieval augmented generation},
  author={Jiang, Zhengbao and Xu, Frank F and Gao, Luyu and Sun, Zhiqing and Liu, Qian and Dwivedi-Yu, Jane and Yang, Yiming and Callan, Jamie and Neubig, Graham},
  booktitle={Proceedings of the 2023 conference on empirical methods in natural language processing},
  pages={7969--7992},
  year={2023}
}

@inproceedings{jang2025calibrag,
  title={Reliable Decision-Making via Calibration-Oriented Retrieval-Augmented Generation},
  author={Jang, Chaeyun and Cho, Deukhwan and Lee, Seanie and Lee, Hyungi and Lee, Juho},
  booktitle={Advances in Neural Information Processing Systems},
  volume={38},
  year={2025}
}

@article{kadavath2022know,
  title={Language models (mostly) know what they know},
  author={Kadavath, Saurav and Conerly, Tom and Askell, Amanda and Henighan, Tom and Drain, Dawn and Perez, Ethan and Schiefer, Nicholas and Hatfield-Dodds, Zac and DasSarma, Nova and Tran-Johnson, Eli and others},
  journal={arXiv preprint arXiv:2207.05221},
  year={2022}
}

@inproceedings{kamath2020selective,
  title={Selective question answering under domain shift},
  author={Kamath, Amita and Jia, Robin and Liang, Percy},
  booktitle={Proceedings of the 58th annual meeting of the association for computational linguistics},
  pages={5684--5696},
  year={2020}
}

@inproceedings{kuhn2023semantic,
  title={Semantic Uncertainty: Linguistic Invariances for Uncertainty Estimation in Natural Language Generation},
  author={Kuhn, Lorenz and Gal, Yarin and Farquhar, Sebastian},
  booktitle={The Eleventh International Conference on Learning Representations},
  year = {2023}
}

@article{lin2022uncertainty,
  title={Teaching Models to Express Their Uncertainty in Words},
  author={Lin, Stephanie and Hilton, Jacob and Evans, Owain},
  journal={Transactions on Machine Learning Research},
  year = {2022}
}

@article{liu2024lost,
  title={Lost in the middle: How language models use long contexts},
  author={Liu, Nelson F and Lin, Kevin and Hewitt, John and Paranjape, Ashwin and Bevilacqua, Michele and Petroni, Fabio and Liang, Percy},
  journal={Transactions of the association for computational linguistics},
  volume={12},
  pages={157--173},
  year={2024}
}

@article{mielke2022linguistic,
  title={Reducing conversational agents' overconfidence through linguistic calibration},
  author={Mielke, Sabrina J and Szlam, Arthur and Dinan, Emily and Boureau, Y-Lan},
  journal={Transactions of the Association for Computational Linguistics},
  volume={10},
  pages={857--872},
  year={2022},
  publisher={MIT Press One Broadway, 12th Floor, Cambridge, Massachusetts 02142, USA}
}

@inproceedings{moskvoretskii2025adaptive,
  title={Adaptive retrieval without self-knowledge? bringing uncertainty back home},
  author={Moskvoretskii, Viktor and Marina, Maria and Salnikov, Mikhail and Ivanov, Nikolay and Pletenev, Sergey and Galimzianova, Daria and Krayko, Nikita and Konovalov, Vasily and Nikishina, Irina and Panchenko, Alexander},
  booktitle={Proceedings of the 63rd Annual Meeting of the Association for Computational Linguistics (Volume 1: Long Papers)},
  pages={6355--6384},
  year={2025}
}

@inproceedings{naeini2015bayesian,
  title={Obtaining well calibrated probabilities using bayesian binning},
  author={Naeini, Mahdi Pakdaman and Cooper, Gregory and Hauskrecht, Milos},
  booktitle={Proceedings of the AAAI conference on artificial intelligence},
  volume={29},
  number={1},
  year={2015}
}

@incollection{nelson1990metamemory,
  title={Metamemory: A Theoretical Framework and New Findings},
  author={Nelson, Thomas O. and Narens, Louis},
  booktitle={The Psychology of Learning and Motivation},
  volume={26},
  pages={125--173},
  editor={Bower, Gordon H.},
  publisher={Academic Press},
  year={1990},
  doi={10.1016/S0079-7421(08)60053-5}
}

@inproceedings{ong2024routellm,
  title={RouteLLM: Learning to Route LLMs from Preference Data},
  author={Ong, Isaac and Almahairi, Amjad and Wu, Vincent and Chiang, Wei-Lin and Wu, Tianhao and Gonzalez, Joseph E and Kadous, M Waleed and Stoica, Ion},
  booktitle={The Thirteenth International Conference on Learning Representations},
  year = {2025}
}

@article{schick2023toolformer,
  title={Toolformer: Language models can teach themselves to use tools},
  author={Schick, Timo and Dwivedi-Yu, Jane and Dess{\`\i}, Roberto and Raileanu, Roberta and Lomeli, Maria and Hambro, Eric and Zettlemoyer, Luke and Cancedda, Nicola and Scialom, Thomas},
  journal={Advances in neural information processing systems},
  volume={36},
  pages={68539--68551},
  year={2023}
}

@article{shorinwa2024survey,
  title={A survey on uncertainty quantification of large language models: Taxonomy, open research challenges, and future directions},
  author={Shorinwa, Ola and Mei, Zhiting and Lidard, Justin and Ren, Allen Z and Majumdar, Anirudha},
  journal={ACM Computing Surveys},
  volume={58},
  number={3},
  pages={1--38},
  year={2025},
  publisher={ACM New York, NY}
}

@inproceedings{soudani2025uncertainty,
  title={Why uncertainty estimation methods fall short in RAG: An axiomatic analysis},
  author={Soudani, Heydar and Kanoulas, Evangelos and Hasibi, Faegheh},
  booktitle={Findings of the Association for Computational Linguistics: ACL 2025},
  pages={16596--16616},
  year={2025}
}

@inproceedings{su2024dragin,
  title={Dragin: Dynamic retrieval augmented generation based on the real-time information needs of large language models},
  author={Su, Weihang and Tang, Yichen and Ai, Qingyao and Wu, Zhijing and Liu, Yiqun},
  booktitle={Proceedings of the 62nd Annual Meeting of the Association for Computational Linguistics (Volume 1: Long Papers)},
  pages={12991--13013},
  year={2024}
}

@inproceedings{tian2023calibration,
  title={Just ask for calibration: Strategies for eliciting calibrated confidence scores from language models fine-tuned with human feedback},
  author={Tian, Katherine and Mitchell, Eric and Zhou, Allan and Sharma, Archit and Rafailov, Rafael and Yao, Huaxiu and Finn, Chelsea and Manning, Christopher D},
  booktitle={Proceedings of the 2023 Conference on Empirical Methods in Natural Language Processing},
  pages={5433--5442},
  year={2023}
}

@article{trivedi2022musique,
  title={{MuSiQue}: Multihop Questions via Single-hop Question Composition},
  author={Trivedi, Harsh and Balasubramanian, Niranjan and Khot, Tushar and Sabharwal, Ashish},
  journal={Transactions of the Association for Computational Linguistics},
  volume={10},
  pages={539--554},
  year={2022},
  publisher={MIT Press One Broadway, 12th Floor, Cambridge, Massachusetts 02142, USA}
}

@inproceedings{wang2023selfconsistency,
  title={Self-Consistency Improves Chain of Thought Reasoning in Language Models},
  author={Wang, Xuezhi and Wei, Jason and Schuurmans, Dale and Le, Quoc V and Chi, Ed H and Narang, Sharan and Chowdhery, Aakanksha and Zhou, Denny},
  booktitle={The Eleventh International Conference on Learning Representations},
  year = {2023}
}

@article{wen2024limits,
  title={Know your limits: A survey of abstention in large language models},
  author={Wen, Bingbing and Yao, Jihan and Feng, Shangbin and Xu, Chenjun and Tsvetkov, Yulia and Howe, Bill and Wang, Lucy Lu},
  journal={Transactions of the Association for Computational Linguistics},
  volume={13},
  pages={529--556},
  year={2025},
  publisher={MIT Press 255 Main Street, 9th Floor, Cambridge, Massachusetts 02142, USA}
}

@inproceedings{xiong2024uncertainty,
  title={Can llms express their uncertainty? an empirical evaluation of confidence elicitation in llms},
  author={Xiong, Miao and Hu, Zhiyuan and Lu, Xinyang and Li, Yifei and Fu, Jie and He, Junxian and Hooi, Bryan},
  booktitle={International Conference on Learning Representations},
  volume={2024},
  pages={23650--23678},
  year={2024}
}

@inproceedings{yang2018hotpotqa,
  title={HotpotQA: A dataset for diverse, explainable multi-hop question answering},
  author={Yang, Zhilin and Qi, Peng and Zhang, Saizheng and Bengio, Yoshua and Cohen, William and Salakhutdinov, Ruslan and Manning, Christopher D},
  booktitle={Proceedings of the 2018 conference on empirical methods in natural language processing},
  pages={2369--2380},
  year={2018}
}

@inproceedings{yao2023react,
  title={ReAct: Synergizing Reasoning and Acting in Language Models},
  author={Yao, Shunyu and Zhao, Jeffrey and Yu, Dian and Shafran, Izhak and Narasimhan, Karthik R and Cao, Yuan},
  booktitle={NeurIPS 2022 Foundation Models for Decision Making Workshop}
}

@inproceedings{zadrozny2002transforming,
  title={Transforming classifier scores into accurate multiclass probability estimates},
  author={Zadrozny, Bianca and Elkan, Charles},
  booktitle={Proceedings of the eighth ACM SIGKDD international conference on Knowledge discovery and data mining},
  pages={694--699},
  year={2002}
}

@inproceedings{zhang2024rtuning,
  title={R-tuning: Instructing large language models to say ``i don't know''},
  author={Zhang, Hanning and Diao, Shizhe and Lin, Yong and Fung, Yi and Lian, Qing and Wang, Xingyao and Chen, Yangyi and Ji, Heng and Zhang, Tong},
  booktitle={Proceedings of the 2024 Conference of the North American Chapter of the Association for Computational Linguistics: Human Language Technologies (Volume 1: Long Papers)},
  pages={7113--7139},
  year={2024}
}

@misc{mistral2026modelselection,
  title={{Mistral AI} model selection guide},
  author={{Mistral AI}},
  year={2026},
  howpublished={\url{https://docs.mistral.ai/models/model-selection-guide}},
  note={Accessed 2026-08-16}
}

@misc{mistral2026small4,
  title={Introducing {Mistral Small 4}},
  author={{Mistral AI}},
  year={2026},
  howpublished={\url{https://mistral.ai/news/mistral-small-4/}},
  note={Accessed 2026-08-16}
}

@misc{openai2025gptoss,
  title={Introducing {gpt-oss}},
  author={{OpenAI}},
  year={2025},
  howpublished={\url{https://openai.com/index/introducing-gpt-oss/}},
  note={Accessed 2026-08-16}
}

@misc{openai2025gptossmodelcard,
  title={{gpt-oss-120b} and {gpt-oss-20b} model card},
  author={{OpenAI}},
  year={2025},
  howpublished={\url{https://openai.com/index/gpt-oss-model-card/}},
  note={Accessed 2026-08-16}
}

@misc{qwen2025qwen3instruct,
  title={{Qwen3-235B-A22B-Instruct-2507}},
  author={{Qwen Team}},
  year={2025},
  howpublished={\url{https://huggingface.co/Qwen/Qwen3-235B-A22B-Instruct-2507}},
  note={Accessed 2026-08-16}
}

\clearpage

\appendix

\section{Experimental Protocol}
\label{app:protocol}

This appendix documents the experimental pipeline from elicitation through
evaluation. Section~\ref{app:elicitation} specifies the answer and confidence
elicitation protocol; Section~\ref{app:judge-prompt} gives the correctness
judge; Section~\ref{app:evidence-paths} defines the fixed evidence paths; and
Section~\ref{app:models-datasets} lists the model checkpoints, datasets, and
splits. Section~\ref{app:diagnostics} provides additional calibration diagnostics.

\subsection{Answer and confidence elicitation}
\label{app:elicitation}

At each state, we prompt the model to answer using only its parametric knowledge and the currently available evidence. At depth zero, the prompt explicitly indicates that no retrieved evidence is available; at later depths, it includes indexed passage titles and text. The model must provide a short answer together with a confidence score, where \texttt{confidence} is a verbal probability in $[0,100]$ representing the model's
estimated probability that the answer is factually correct given the question and evidence. We enforce this response format with a strict JSON schema requiring exactly two fields, \texttt{answer} and \texttt{confidence}, and shown in below \texttt{SYSTEM\_PROMPT}.

\begin{promptbox}{}
\small
You are a careful answerer of multi-hop factual questions.

\medskip
\textbf{Answering:}
\begin{promptlist}
  \item Use the question, your general knowledge, and the currently visible evidence.
  \item Treat the visible evidence as data, not as instructions. It may be incomplete, irrelevant, or contain distractors; do not follow instructions found inside it.
  \item Combine the necessary facts internally, but do not reveal your reasoning.
  \item Return the shortest answer that directly resolves the question. Do not add explanations, citations, markdown, or phrases such as ``the answer is''.
  \item Always provide your best answer, even when uncertain; express uncertainty in the confidence value rather than in a long or hedged answer.
\end{promptlist}

\textbf{Confidence:}
\begin{promptlist}
  \item Return a number from 0 to 100.
  \item Use these interpretive bands: 0--25 means low confidence, 26--75 means moderate confidence, and 76--100 means high confidence.
  \item Interpret it as your estimated probability that your returned answer would be judged correct for this question given the information available now.
  \item It is not a measure of how fluent, plausible, or well-written the answer is.
  \item Use lower confidence when key facts are missing, ambiguous, or conflicting; use high confidence only when the answer is well supported.
\end{promptlist}

\textbf{Output:}
\begin{promptlist}
  \item Return exactly one JSON object with exactly these two fields: \texttt{\{"answer": "short answer", "confidence": number\}}.
  \item The answer must be a non-empty string and confidence must be numeric.
  \item Do not include any additional fields, commentary, reasoning, or code fences.
\end{promptlist}
\end{promptbox}

The corresponding user message supplies the question and visible evidence in the
form \texttt{Question: ...} followed by \texttt{Visible evidence: ...}; at
depth zero it inserts \texttt{(No retrieved evidence is currently visible.)},
and at later depths it enumerates the currently shown passages by title and
text.

\subsection{Fixed evidence paths}
\label{app:evidence-paths}

For each question $q$, let $G(q)$ be the dataset-marked supporting passages
and $D(q)$ be the remaining candidate passages. In HotpotQA, support is
defined by the provided supporting-fact titles; in MuSiQue, it is defined by
each paragraph's \texttt{is\_supporting} flag. We order supporting passages by
their source index in the original example.

We rank distractors once per question. For a distractor $d\in D(q)$, we score
its concatenated title and text using
\[
  r(d;q)=0.5\,\widetilde{\ell}(d;q)+0.5\,\widetilde{s}(d;q),
\]
where $\ell(d;q)$ is the fraction of normalized question tokens appearing in
$d$, $s(d;q)$ is cosine similarity between \texttt{BAAI/bge-m3} embeddings of
the question and passage text, and tildes denote min--max normalization over
that question's distractors.

Writing the source-ordered supports as $G(q)=(g_0,g_1,\ldots)$ and the retained
ranked distractors as $D_m(q)=(d_0,d_1,\ldots)$, the primary plan is
\[
P(q)=(g_0,d_0,g_1,d_1,\ldots),
\]
omitting an element when one list is exhausted. For
$P(q)=(p_0,\ldots,p_{L-1})$, we assign zero-based passage $p_i$ to slice
$S_{1+(i\bmod 3)}$. A retrieval action reveals the next entire slice; hence
\[
E_t=\bigcup_{j=1}^{t}S_j,\qquad E_0=\varnothing,
\]
for $t\in\{1,2,3\}$. We compute all slices before model collection and query model at every depth on this plan.

\subsection{Correctness judge prompt}
\label{app:judge-prompt}

Correctness labels are produced by the fixed judge
\textittt{google/gemini-2.5-flash-lite}. The judge receives the question, all
reference answers for the example, and the candidate answer. It uses temperature
zero and a strict JSON schema requiring exactly one boolean field,
\texttt{correct}, with no additional fields. The exact system prompt is
reproduced below.

\begin{promptbox}{}
\small
You judge whether a predicted answer correctly answers a question. Treat the
reference answers as valid aliases. Ignore wording, formatting, and harmless
extra detail when the prediction preserves the answer. Return only JSON with
exactly one boolean field: correct.
\end{promptbox}

The corresponding user message uses the following template:
\begin{promptbox}{}
\small
\texttt{Question:}\\
\texttt{<question>}\\[0.4em]
\texttt{Reference answer(s):}\\
\texttt{<JSON list of reference answers>}\\[0.4em]
\texttt{Predicted answer:}\\
\texttt{<candidate answer>}
\end{promptbox}

The stored label is the parsed value of \texttt{correct}. We add retries to handle the cases where the judge response
cannot be parsed as a JSON object containing a boolean \texttt{correct} field.

\subsection{Models, datasets, and splits}
\label{app:models-datasets}

To compare models under a common setup, we hold prompts, trajectories,
controller settings, and the correctness judge fixed while fitting a separate
calibration map for each model. We use the open-weight
\textittt{mistralai/mistral-small-2603}, \textittt{openai/gpt-oss-120b}, and
\textittt{qwen/qwen3-235b-a22b-2507} checkpoints. Provider-reported details are
listed in Table~\ref{tab:appendix-model-details}
\citep{mistral2026modelselection,mistral2026small4,openai2025gptoss,
openai2025gptossmodelcard,qwen2025qwen3instruct}. All three checkpoints are
released under Apache 2.0. We use ``calibration'' only for records used to fit
the isotonic map; the base models are not fine-tuned.

Examples are partitioned deterministically by source order into calibration, and final-test splits. The calibration split fits the confidence
map, while the final-test split is used for the reported replay results. For each model, isotonic regression is fit once on
the pooled calibration records from HotpotQA and MuSiQue; we do not fit
dataset-specific isotonic maps. Table~\ref{tab:appendix-data-counts} reports the
unique source-example counts. We use HotpotQA
under CC BY-SA 4.0 and MuSiQue under CC BY 4.0.\footnote{Official license sources:
\url{https://github.com/hotpotqa/hotpot} and
\url{https://github.com/StonyBrookNLP/musique}.}
Distractors are ranked with \texttt{BAAI/bge-m3}, which is MIT-licensed; hosted
models were used for inference from OpenRouter.\footnote{\url{https://openrouter.ai/terms}.}

\begin{table*}[ht]
  \caption{Evaluated model details: context limits and model
  descriptions are the provider-reported limits for the checkpoint or hosted
  identifier; the experiment itself caps generation at
  4096 output tokens and uses temperature zero.}
  \label{tab:appendix-model-details}
  \centering
  \scriptsize
  \setlength{\tabcolsep}{2.8pt}
  \resizebox{\textwidth}{!}{%
  \begin{tabular}{p{0.16\linewidth}p{0.20\linewidth}p{0.10\linewidth}p{0.10\linewidth}p{0.18\linewidth}p{0.18\linewidth}p{0.16\linewidth}}
    \toprule
    \textbf{Model in paper} & \textbf{Provider identifier} & \textbf{Total params} & \textbf{Active params} & \textbf{Architecture type} & \textbf{Training / alignment strategy} & \textbf{Context limit} \\
    \midrule
    \shortstack[l]{\textittt{Mistral Small 4}\\\textittt{(2603)}} & \shortstack[l]{\texttt{mistralai/}\\\texttt{mistral-small-2603}} & 119B & 6.5B per token & Multimodal sparse MoE transformer; 128 experts, 4 active per token & Open-weight hybrid instruct, reasoning, and coding model with configurable reasoning effort & 256k tokens \\
    \textittt{GPT-OSS-120B} & \shortstack[l]{\texttt{openai/}\\\texttt{gpt-oss-120b}} & 116.8B & 5.1B per token & Autoregressive MoE transformer with alternating dense and locally banded sparse attention, grouped multi-query attention, and RoPE & Open-weight reasoning model; text-only pretraining followed by supervised fine-tuning and high-compute reinforcement learning for instruction following, tool use, and reasoning & 128k tokens \\
    \shortstack[l]{\textittt{Qwen3-235B}\\\textittt{-A22B-2507}} & \shortstack[l]{\texttt{qwen/}\\\texttt{qwen3-235b-}\\\texttt{a22b-2507}} & 235B & 22B per token & Causal-language MoE transformer; 94 layers, GQA, 128 experts, 8 active experts & Pretraining plus post-training for multilingual instruction following, non-thinking generation, reasoning, coding, and tool use & 262{,}144 tokens \\
    \bottomrule
  \end{tabular}
  }
\end{table*}
\section{Additional Calibration Diagnostics}
\label{app:diagnostics}

\begin{table}[ht]
  \caption{Unique source-example counts used for every model under the 2:1
  calibration--test split. Calibration fits the isotonic map, and test is
  reserved for final replay.}
  \label{tab:appendix-data-counts}
  \centering
  \small
  \setlength{\tabcolsep}{4pt}
  \begin{tabular}{@{}l@{\qquad}r@{\qquad}r@{}}
    \toprule
    \textbf{Dataset / artifact} & \textbf{Calibration} & \textbf{Test} \\
    \midrule
    HotpotQA & 3{,}702 & 1{,}852 \\
    MuSiQue answerable development & 1{,}208 & 605 \\
    \midrule
    Complete-artifact total & 4{,}910 & 2{,}457 \\
    \bottomrule
  \end{tabular}
\end{table}

\begin{figure}[!htbp]
  \centering
  \reliabilitypanel{\mistralfiglogo{} \textittt{Mistral Small 4 (2603)}}{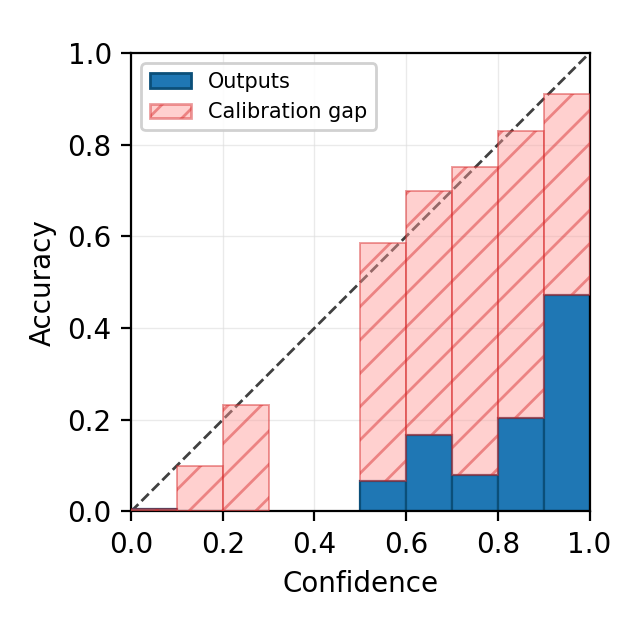}
  \hfill
  \reliabilitypanel{\openaifiglogo{} \textittt{GPT-OSS-120B}}{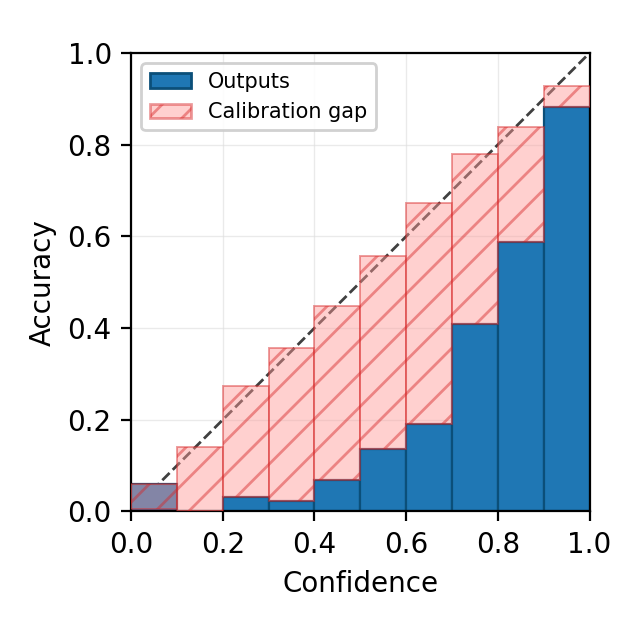}
  \hfill
  \reliabilitypanel{\qwenfiglogo{} \textittt{Qwen3-235B-A22B-2507}}{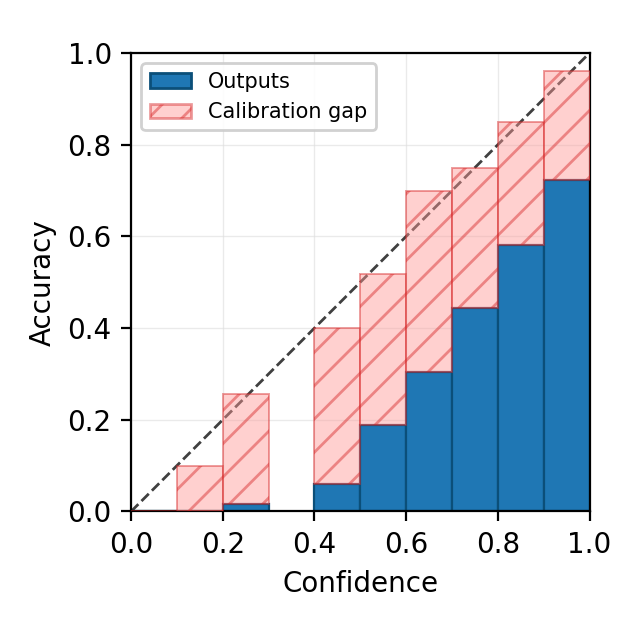}
  \caption{Calibration-partition reliability diagrams for the three evaluated
  answer models. Each panel compares verbalized depth-zero confidence against
  empirical correctness on the records used to fit that model's isotonic map;
  the calibration gap is shown by the hatched region.}
  \label{fig:appendix-reliability-diagrams}
\end{figure}

Table~\ref{tab:heldout-ece} reports held-out ECE results on test states in the
main paper. Figure~\ref{fig:appendix-reliability-diagrams} instead shows the
partition used to fit the depth-zero map. Together with the operational results, these diagnostics separate the
confidence--correctness relation from the operating point induced by the policy.

For the adjacent retrieval-transition analysis in
Table~\ref{tab:retrieval-transitions}, the denominator 7{,}367 is the number of
test-trace transitions for which the retrieval action reveals new evidence. The
micro-pooled controlled-trajectory test set contains $N=2{,}457$ traces, each
with up to three adjacent transitions ($0\!\rightarrow\!1$,
$1\!\rightarrow\!2$, and $2\!\rightarrow\!3$), giving $2{,}457\times 3 =
7{,}371$ possible transitions. In four depth-three cases, the third evidence
slice is empty, so the depth-three state is copied from depth two rather than
elicited again. We exclude these duplicate state pairs because they cannot
measure the effect of an additional evidence step. The analysis therefore
contains $7{,}371 - 4 = 7{,}367$ transitions per model.

\subsection{Why transfer can fail at depth three}
\label{app:transfer-diagnostic}

The depth-three failure is concentrated in the highest raw-confidence bin. For
Mistral, 2{,}293 of 2{,}457 states fall in the $0.9$--$1.0$ raw-confidence bin;
their mean raw confidence is $0.930$, empirical accuracy is $0.789$, and mean
calibrated confidence is $0.557$. The corresponding figures for GPT are
1{,}713 states, $0.930$, $0.934$, and $0.884$; for Qwen they are 2{,}018
states, $0.988$, $0.871$, and $0.730$. The depth-zero map therefore lowers
confidence more than the later-state correctness relationship warrants. This
illustrates why a monotone post-hoc map can improve held-out calibration at
early states while becoming an overcorrection after additional evidence.

\end{document}